\documentclass{ieeeaccess}
\usepackage{cite}
\usepackage{amsmath,amssymb,amsfonts}
\usepackage{algorithmic}
\usepackage{graphicx}
\usepackage{textcomp}

\usepackage{url}
\usepackage{makecell}
\usepackage{float}
\usepackage{booktabs}
\usepackage{multirow}
\usepackage{array}
\usepackage{pifont}    
\usepackage{arydshln}

\newcommand{\cmark}{\ding{51}}%
\newcommand{\xmark}{\ding{55}}%
\newcommand{\nr}{\textcolor[gray]{0.5}{\textit{n/r}}}

\usepackage[shortcuts,acronym]{glossaries} 
\usepackage[shortcuts,acronym]{glossaries}

\newacronym{uwb}{UWB}{ultra-wideband}
\newacronym{ir-uwb}{IR-UWB}{impulse-radio ultra-wideband}
\newacronym{fmcw}{FMCW}{frequency-modulated continuous wave}
\newacronym{ble}{BLE}{Bluetooth low energy}
\newacronym{mmwave}{mmWave}{millimeterwave}
\newacronym{gnss}{GNSS}{global navigation satellite system}
\newacronym{elu}{ELU}{exponential linear unit}
\newacronym{fr3}{FR3}{frequency range 3}
\newacronym{ch5}{CH5}{channel 5}
\newacronym{ch9}{CH9}{channel 9}
\newacronym{jcas}{JCAS}{joint communication and sensing}
\newacronym{isac}{ISAC}{integrated sensing and communication}
 
\newacronym{ioe}{IoE}{Internet of Everything}
\newacronym{iot}{IoT}{Internet of Things}
\newacronym{iiot}{IIoT}{Industrial Internet of Things}
 
\newacronym{psd}{PSD}{power spectral density}

\newacronym{afsiw}{AFSIW}{air-filled substrate-integrated-waveguide}
\newacronym{cbs}{CBS}{cavity-backed slot}
\newacronym{dra}{DRA}{dielectric resonator antenna}
\newacronym{mimo}{MIMO}{multiple input multiple output}
\newacronym{hpbw}{HPBW}{half-power beamwidth}
\newacronym{ftbr}{FTBR}{front-to-back ratio}
\newacronym{sff}{SFF}{system fidelity factor}
\newacronym{dee}{DEE}{distance estimation error}
\newacronym{fov}{FOV}{field-of-view}
\newacronym{qm}{QM}{quarter-mode}

\newacronym{rtls}{RTLS}{real-time location system}
\newacronym{tof}{ToF}{time of flight}
\newacronym{rtt}{RTT}{round trip time}
\newacronym{toa}{ToA}{time of arrival}
\newacronym{twr}{TWR}{two-way ranging}
\newacronym{tdoa}{TDoA}{time difference of arrival}
\newacronym{pdoa}{PDoA}{phase difference of arrival}
\newacronym{aoa}{AoA}{angle of arrival}
\newacronym{acir}{ACIR}{accumulated channel impulse response}
\newacronym{cir}{CIR}{channel impulse response}
\newacronym{csi}{CSI}{channel state information}
\newacronym{dop}{DOP}{dilution of precision}
 
\newacronym{mae}{MAE}{mean absolute error}
\newacronym{rmse}{RMSE}{root mean square error}

\newacronym{music}{MUSIC}{Multiple Signal Classification}
\newacronym{fft}{FFT}{fast Fourier transform}
\newacronym{stft}{STFT}{short-time Fourier transform}
\newacronym{ekf}{EKF}{extended Kalman filter}
\newacronym{iq}{IQ}{in-phase and quadrature}
\newacronym{snr}{SNR}{signal-to-noise ratio}
 
\newacronym{tdm}{TDM}{time Doppler map}
\newacronym{rdm}{RDM}{range Doppler map}

\newacronym{gpu}{GPU}{graphics processing unit}
\newacronym{fpu}{FPU}{floating-point unit}
\newacronym{dat}{DAT}{dual-antenna tile}
\newacronym{cots}{COTS}{commercial off-the-shelf}
\newacronym{trx}{TRX}{transceiver}
\newacronym{ic}{IC}{integrated circuit}
\newacronym{pcb}{PCB}{printed circuit board}
\newacronym{tcxo}{TCXO}{temperature-controlled crystal oscillator}
\newacronym{pu}{PU}{processing unit}
\newacronym{pmic}{PMIC}{power management integrated circuit}
 
\newacronym{spi}{SPI}{Serial Peripheral Interface}
\newacronym{i2c}{I2C}{Inter-Integrated Circuit}
\newacronym{csn}{CSn}{Chip Select}
\newacronym{sda}{SDA}{Serial Data}
\newacronym{scl}{SCL}{Serial Clock}
\newacronym{gpio}{GPIO}{general-purpose input-output}
\newacronym{io}{IO}{input-output}
 
\newacronym{rtos}{RTOS}{real-time operating system}
\newacronym{soc}{SoC}{system-on-chip}
\newacronym{rf}{RF}{radio frequency}
\newacronym{gcpw}{GCPW}{grounded coplanar waveguide}
\newacronym{tx}{TX}{transmit}
\newacronym{rx}{RX}{receive}

\newacronym{los}{LOS}{line-of-sight}
\newacronym{nlos}{NLOS}{non-line-of-sight}
 
\newacronym{roi}{ROI}{region of interest}
 
\newacronym{dk}{DK}{development kit}
\newacronym{sram}{SRAM}{static random access memory}
 
\newacronym{fpga}{FPGA}{field-programmable gate array}
 
\newacronym{sdr}{SDR}{software defined radio}
 
\newacronym{fp}{FP}{first path}

\newacronym{har}{HAR}{human activity recognition}
\newacronym{ml}{ML}{machine learning}
\newacronym{ai}{AI}{artificial intelligence}
\newacronym{cnn}{CNN}{convolutional neural network}
\newacronym{lstm}{LSTM}{long short-term memory}
\newacronym{dnn}{DNN}{deep neural network}
\newacronym{nn}{NN}{neural network}
 
\newacronym{rr}{RR}{respiration rate}
\newacronym{br}{BR}{breathing rate}
\newacronym{hr}{HR}{heart rate}
\newacronym{rbm}{RBM}{random body movement}
\newacronym{bpm}{BPM}{breaths per minute}
 
\newacronym{pll}{PLL}{phase locked loop}
\newacronym{shr}{SHR}{synchronization header}
\newacronym{phr}{PHR}{physical layer header}
\newacronym{psdu}{PSDU}{physical layer service data unit}

\newacronym{mac}{MAC}{medium access control}
\newacronym{rs}{RS}{Reed-Solomon}
 
\newacronym{tts}{TTS}{text-to-speech}
 
\newacronym{lopo}{LOPO}{leave-one-person-out}
\newacronym{loso}{LOSO}{leave-one-scenario-out}
\newacronym{lo2so}{LO2SO}{leave-two-scenarios-out}
\newacronym{lobpo}{LOBPO}{leave-one-bed-position-out}
 
\newacronym{ofdm}{OFDM}{orthogonal frequency division multiplexing}
\usepackage{cleveref} 
\usepackage{soul} 
\crefname{table}{table}{tables}
\Crefname{table}{Table}{Tables}
\crefname{figure}{fig.}{Figures}
\Crefname{figure}{Fig.}{Figures}
\Crefname{section}{Section}{Sections}

\usepackage{bm}
\makeatletter
\AtBeginDocument{\DeclareMathVersion{bold}
\SetSymbolFont{operators}{bold}{T1}{times}{b}{n}
\SetSymbolFont{NewLetters}{bold}{T1}{times}{b}{it}
\SetMathAlphabet{\mathrm}{bold}{T1}{times}{b}{n}
\SetMathAlphabet{\mathit}{bold}{T1}{times}{b}{it}
\SetMathAlphabet{\mathbf}{bold}{T1}{times}{b}{n}
\SetMathAlphabet{\mathtt}{bold}{OT1}{pcr}{b}{n}
\SetSymbolFont{symbols}{bold}{OMS}{cmsy}{b}{n}
\renewcommand\boldmath{\@nomath\boldmath\mathversion{bold}}}
\makeatother

\def\BibTeX{{\rm B\kern-.05em{\sc i\kern-.025em b}\kern-.08em
    T\kern-.1667em\lower.7ex\hbox{E}\kern-.125emX}}

\begin{document}
\history{Date of publication xxxx 00, 0000, date of current version xxxx 00, 0000.}
\doi{PLACEHOLDER}

\title{A comparison between ceiling-mounted FMCW, IR-UWB and Wi-Fi radar for in-bedroom human activity monitoring and sleep interruption detection}
\author{\uppercase{Anton Lambrecht}\authorrefmark{1},
\uppercase{Reda El Hail}\authorrefmark{2,3}, \uppercase{Xianjun Jiao}\authorrefmark{1},
\uppercase{Pieter Crombez}\authorrefmark{4},
\uppercase{Dominique Schreurs}\authorrefmark{5}, 
\uppercase{Peter Karsmakers}\authorrefmark{2,3}, \uppercase{Adnan Shahid}\authorrefmark{1}, AND \uppercase{Eli De Poorter}\authorrefmark{1}
}

\address[1]{IDLab, Department of Information Technology, Ghent University-imec, iGent Tower, Technologiepark-Zwijnaarde 126, B-9052 Ghent, Belgium. (email: Anton.Lambrecht@Ugent.be)}
\address[2]{Department of Computer Science, Leuven AI, KU Leuven, B-2440 Geel, Belgium (e-mail: Reda.Elhail@kuleuven.be)}
\address[3]{Flanders Make, MPRO, B-3000 Leuven, Belgium (e-mail: Reda.Elhail@kuleuven.be)}
\address[4]{Televic Healthcare, 8870 Izegem, Belgium}
\address[5]{Div. ESAT-WAVECORE, KU Leuven, Belgium (e-mail:dominique.schreurs@kuleuven.be)}

\tfootnote{The research that led to these results has received funding from the DistriMuSe project (HORIZON-KDT-JU-2023-2-RIA) with Grant No 101139769, the CORNET project DARTS (HBC.2023.0772) and from the NAV-ALERT project from the Belgian Defense through contract number 24DEFRA009.}

\markboth
{Lambrecht \headeretal: A comparison between ceiling-mounted FMCW, IR-UWB and Wi-Fi radar}
{Lambrecht \headeretal: A comparison between ceiling-mounted FMCW, IR-UWB and Wi-Fi radar}

\corresp{Corresponding author: Anton Lambrecht (email: Anton.Lambrecht@Ugent.be).}

\begin{abstract}
Despite their growing importance for contact-free radio frequency (RF) based healthcare monitoring, different radio technologies such as frequency-modulated continuous wave (FMCW) radar, impulse radio ultra-wideband (IR-UWB), and Wi-Fi sensing are rarely compared under identical deployment conditions, as existing studies typically differ in hardware, datasets, and evaluation methodologies. In addition, the performance of ceiling-mounted radars, despite their practical deployment and cost advantages in healthcare environments, remain underexplored. Therefore, this paper presents a controlled comparison and analysis of ceiling-mounted FMCW, IR-UWB, and Wi-Fi sensing using synchronized recordings from 20 participants across six room layouts. All technologies are evaluated with the same convolutional neural network (CNN) on both a fine-grained 10-class human activity recognition (HAR) task and a coarse 4-class sleep monitoring task. IR-UWB achieves the highest cross-subject activity recognition performance (89.0\% macro F1), while FMCW generalizes best to unseen room layouts (83.8\% macro F1). For sleep monitoring, all technologies exceed 92\% macro F1 in unseen environments. The results reveal a fundamental trade-off between recognition performance and environmental robustness, which can be explained through differences in range resolution, antenna diversity, Doppler resolution, and spatial information retention. These findings provide practical guidelines for the design of healthcare-oriented RF sensing systems.
\end{abstract}


\titlepgskip=-21pt

\maketitle

\section{Introduction}
\label{sec:introduction}

\PARstart{W}{ith} the growing demand for non-intrusive monitoring in healthcare and assisted living, contactless \ac{har} is emerging as a promising approach for monitoring patient behavior and assessing sleep quality in both domestic and clinical settings. By 2050, the global population aged over 60 is projected to reach 2.1~billion, intensifying the need for scalable and privacy-preserving life-quality monitoring solutions~\cite{who_ageing_2025}. Sleep disorders are particularly prevalent among older adults: a meta-analysis spanning 36 countries found that 40\% experience poor sleep quality and 29\% suffer from insomnia~\cite{canever_worldwide_2024}. These conditions are strongly associated with increased fall risk, cognitive decline, dementia onset, and reduced quality of life. Reliable monitoring of human activity and sleep behavior can enable early detection of health deterioration, support timely intervention, and improve patient safety and quality of care.

Current monitoring systems often rely on camera-based approaches, which raise significant privacy and acceptability concerns in sensitive environments. A survey of 304 informal caregivers of people with dementia showed that \ac{rf} sensing systems are rated more acceptable (4.0/5) than camera-based systems (3.1/5)~\cite{wrede_understanding_2023}, motivating the use of contactless \ac{rf}-based sensing as a privacy-preserving alternative. Among \ac{rf}-based technologies, \ac{ir-uwb}, \ac{fmcw} radar, and Wi-Fi represent three prominent approaches with distinct design trade-offs. Wi-Fi is widely deployed and offers the potential for sensing with minimal additional infrastructure, but is constrained by limited bandwidth. In contrast, \ac{fmcw} is a mature sensing-first modality with limited communication integration, while \ac{ir-uwb} occupies a middle ground, offering higher range resolution than Wi-Fi while maintaining full communication capabilities with IEEE-compliant \ac{cots} hardware.

This paper presents a comparative study of \ac{ml}-based \ac{har} and sleep monitoring across three \ac{cots} sensing technologies: \ac{fmcw} radar, \ac{ir-uwb}, and Wi-Fi. We evaluate performance using both a fine-grained activity label set for general healthcare use cases and a coarser sleep-specific label set. A uniform \ac{cnn} architecture is employed to ensure that observed performance differences reflect signal characteristics rather than model design. This enables a fair assessment of each modality’s suitability for reliable, privacy-preserving \ac{har}.

The contributions of this paper are:
\begin{enumerate}
\item A controlled comparison of \ac{fmcw}, \ac{ir-uwb}, and Wi-Fi sensing using synchronized recordings acquired simultaneously in an identical ceiling-mounted deployment. All modalities are processed using the same learning pipeline and evaluated under multiple cross-validation protocols, allowing the observed differences to be attributed to the sensing technology rather than the experimental design.
\item A signal-level analysis linking sensing properties such as range resolution, Doppler resolution, antenna diversity, and spatial information retention to \ac{har} performance and robustness. Based on these insights, practical guidelines are derived for selecting suitable signal representations and preprocessing strategies for different \ac{rf} sensing technologies.
\item An open synchronized dataset containing \ac{fmcw}, \ac{ir-uwb}, and Wi-Fi measurements from 20 participants across six room layouts, enabling reproducible benchmarking and future research on cross-modality \ac{rf} sensing. \footnote{https://gitlab.ilabt.imec.be/datasets/Activity-recognition-datasets}
\end{enumerate}

\section{Related works}
 
\begin{table*}[!t]
\caption{Comparison of representative RF-based human activity recognition studies in terms of sensing hardware, deployment setup, and dataset characteristics. In contrast to previous work, this study compares FMCW, IR-UWB, and Wi-Fi using synchronized recordings acquired simultaneously in an identical ceiling-mounted setup, enabling a controlled comparison across sensing modalities. \cmark/\xmark{} indicate presence/absence of the property. \nr{} = not reported. \textsuperscript{\dag}~ In this paper, the Wi-Fi transmitter is ceiling-mounted while the receiver is placed on the floor.}
\label{tab:related_work}
\centering
\footnotesize
\setlength{\tabcolsep}{5pt}
\renewcommand{\arraystretch}{1.15}
\begin{tabular*}{\textwidth}{@{\extracolsep{\fill}} l c l l l c c c c c @{}}
\toprule
 & & \multicolumn{3}{c}{\textbf{Hardware}} & \multicolumn{5}{c}{\textbf{Dataset}} \\
\cmidrule(lr){3-5} \cmidrule(l){6-10}
\textbf{Ref.} & \textbf{Year} & \textbf{Type} & \textbf{Device} & \textbf{Setup}
 & \textbf{\#ppl} & \textbf{\#rooms} & \makecell{\textbf{Ceiling}\\\textbf{mount}}
 & \textbf{\#lbl} & \makecell{\textbf{Nat.}\\\textbf{flow}} \\
\midrule
\cite{lu_unobtrusive_2023}      & 2023 & IR-UWB & X4M02 & monostatic & 10 & 2 & \cmark & 2  & \xmark \\ \hdashline
\cite{lu_office_2025}           & 2025 & IR-UWB & P440  & monostatic & 10  & 1 & \cmark & 5  & \xmark \\ \hdashline
\cite{guo_human_2024} & 2024 & \makecell[l]{Wi-Fi}
  & \makecell[l]{Intel 5300}
  & \makecell[l]{bi-static}
  & 5 & 1 & \xmark & 8 & \cmark \\ \hdashline
\cite{chen_wiphase_2025} & 2025 & Wi-Fi
  & \makecell[l]{Intel 5300}
  & bi-static
  & 10 & 2 & \xmark & 6 & \xmark \\ \hdashline
\makecell[l]{\cite{gurbuz_cross-frequency_2020, ding_radar_2025, ibrahim_radspecfusion_2025}} & \makecell[l]{2020, 2025} & \makecell[l]{IR-UWB, FMCW}
  & \makecell[l]{Xethru X4\\Ancortek SDR-Kit\\TI IWR1443}
  & monostatic & 6 & \nr & \xmark & 11 & \xmark \\ \hdashline
\cite{chen_rf-based_2023}          & 2023 & \makecell[l]{IR-UWB\\Wi-Fi}
  & \makecell[l]{Xethru\\Intel 5300}
  & \makecell[l]{monostatic\\bi-static}
  & \makecell[c]{30\\ \nr} & \makecell[c]{7\\ \nr} & \makecell[c]{\cmark\\ \xmark\textsuperscript{\dag}} & 7 & \xmark \\ \hdashline
\cite{dahal_comparison_2024}          & 2024 & \makecell[l]{FMCW\\Wi-Fi}
  & \makecell[l]{Radarbook2\\RPi 3B+ (Nexmon)}
  & \makecell[l]{monostatic\\bi-static}
  & 5 & 1 & \xmark & 7 & \xmark \\
\midrule
\makecell[l]{\textbf{This} \textbf{work}} &  & \makecell[l]{FMCW\\IR-UWB\\Wi-Fi}
  & \makecell[l]{IWR6843AOP\\DW3000\\ZedBoard/openwifi}
  & \makecell[l]{monostatic\\pseudo monostatic\\monostatic}
  & 20 & 6 & \cmark & 10 & \cmark \\
\bottomrule
\end{tabular*}
\end{table*}

Table~\ref{tab:related_work} summarizes representative RF-based \ac{har} studies. Although numerous works have investigated \ac{fmcw} radar, \ac{ir-uwb}, or Wi-Fi individually, controlled comparisons between sensing modalities in identical conditions remain scarce. Existing comparative studies typically differ in sensing hardware, deployment geometry, dataset design, or evaluation methodology, making it difficult to attribute observed performance differences to the sensing modality itself.

Most prior work focuses on a single sensing technology. \ac{ir-uwb} studies~\cite{lu_unobtrusive_2023, lu_office_2025} predominantly employ sensing-oriented, non-IEEE-compliant hardware in monostatic configurations, whereas Wi-Fi-based approaches~\cite{guo_human_2024, chen_wiphase_2025, chen_rf-based_2023} almost exclusively rely on bi-static deployments. This is largely because Wi-Fi sensing studies typically leverage off-the-shelf commodity devices, where access to the transmitted and received signals is distributed across separate nodes. In contrast, SDR-based implementations such as openwifi provide full control over the physical layer and enable signal transmission and reception on the same platform, thereby facilitating monostatic radar operation. Therefore, although monostatic Wi-Fi radar has been explored, existing work has primarily focused on respiration-rate determination \cite{jiao_single-input-multiple-output_2025, kristensen_sdr-based_2025, bakhshi_respisense_2025}, rather than on \ac{har}.

Another important distinction is deployment geometry. Most prior work adopts side-looking or wall-mounted configurations, while ceiling-mounted systems remain comparatively rare. Although side-looking deployments (\cite{guo_human_2024, chen_wiphase_2025, gurbuz_cross-frequency_2020, ding_radar_2025, ibrahim_radspecfusion_2025, dahal_comparison_2024}) maximize the human radar cross-section and simplify the sensing task, ceiling-mounted sensing is considerably easier to integrate non-intrusively into real-world healthcare environments and this under-explored research area is therefore the focus of this work.

Thirdly, cross-modality studies remain limited and do not isolate the sensing modality as the primary experimental variable. Only a few works compare different radar technologies to study domain adaptation or modality fusion~\cite{gurbuz_cross-frequency_2020, ding_radar_2025, ibrahim_radspecfusion_2025}. These studies evaluate sensing-first radar platforms (e.g., custom IR-UWB systems and millimeter-wave radar) that share similar sensing principles and are generally not IEEE-compliant communication systems. While these works demonstrate transferability or complementarity across radar modalities, they do not provide a controlled comparison of the sensing technologies themselves. Other studies compare \ac{ir-uwb} with Wi-Fi~\cite{chen_rf-based_2023} or \ac{fmcw} with Wi-Fi~\cite{dahal_comparison_2024}, but both pair a monostatic radar with a spatially separated Wi-Fi link, so the geometry changes together with the sensing technology. Chen et al. evaluate \ac{ir-uwb} with more than 30 participants across seven environments, although the characteristics of the Wi-Fi measurements are not separately reported. Dahal et al. evaluate both modalities in a single controlled laboratory environment with five participants.


Experimental design further complicates comparison. Existing datasets are generally limited in participant or environmental diversity and commonly record activities as repeated isolated actions rather than continuous activity sequences. While repeated executions simplify recognition, they remove the transitions and execution variability encountered in practical deployments, making the resulting performance less representative of real-world use.

In contrast, this work isolates the sensing modality as the primary experimental variable by evaluating \ac{fmcw}, \ac{ir-uwb}, and Wi-Fi using synchronized recordings acquired simultaneously in an identical ceiling-mounted configuration (\Cref{fig:system_overview}). All modalities are processed using the same learning pipeline and evaluated under identical protocols on a dataset comprising 20 participants, six room layouts, and natural activity flows. 

\section{System model}

\begin{figure*}[!t]
    \centering\includegraphics[width=0.95\linewidth]{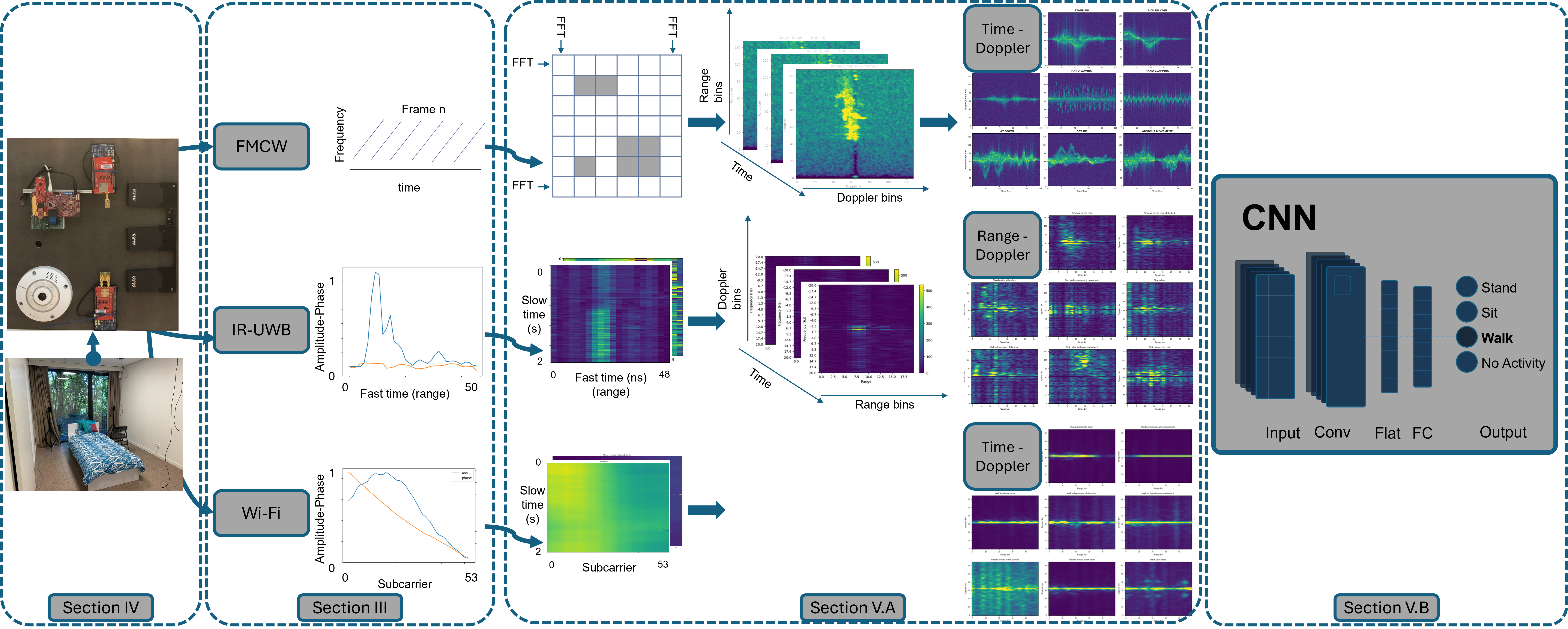}
\caption{Complete system overview from data acquisition to activity classification. The pipeline illustrates the physical measurement setup (left), the raw signal acquisition and mathematical models for \ac{fmcw}, \ac{ir-uwb}, and Wi-Fi (center-left), the modality-specific preprocessing into Time-Doppler and Range-Doppler representations (center-right), and the uniform Convolutional Neural Network (CNN) architecture used to predict the final activity classes.}
    \label{fig:system_overview}
\end{figure*}

As shown in \Cref{fig:system_overview}, \ac{fmcw}, \ac{ir-uwb} and Wi-Fi capture time-varying channel reflections through different representations: beat signals, \ac{cir} and \ac{csi}, respectively. Despite these measurement differences, all three can be described within a common system-level framework, which is introduced first and subsequently specialized for each technology.

\subsection{Problem statement}
\label{sec:system_model_general}

All three sensing modalities observe human motion through time-varying reflections of the wireless propagation channel and can therefore be modeled as different measurements of the same dynamic multipath environment. \Ac{fmcw} probes the channel with chirps, \ac{ir-uwb} with pulses and Wi-Fi with \ac{ofdm} packets from which per-subcarrier \ac{csi} is extracted. Stacking successive complex-valued channel observations yields the measurement matrix
\begin{equation}
    X^{(j)} = x^{(j)}[m,k],
    \qquad
    1 \leq m \leq M_j,\;
    1 \leq k \leq K_j,
    \label{eq:general_matrix}
\end{equation}
where $j \in \{\text{FMCW},\, \text{IR-UWB},\, \text{Wi-Fi}\}$ denotes the sensing technology. The row index $m$ represents slow-time (successive chirps, pulses, or packets), with
\begin{equation}
    M_j = w \cdot sr_j ,
    \label{eq:slow_time_rows}
\end{equation}
for an observation window of duration $w$ and slow-time sampling rate $sr_j$. The column index $k$ spans a modality-dependent fast dimension, whose physical meaning and size $K_j$ follow from the acquisition principle:
\begin{equation}
    K_j =
    \begin{cases}
        N_s, & j = \text{FMCW} \quad \text{(beat-signal samples per chirp)},\\[4pt]
        N_f, & j = \text{IR-UWB} \quad \text{(fast-time delay bins)},\\[4pt]
        N_c, & j = \text{Wi-Fi} \quad \text{(\ac{ofdm} subcarriers)}.
    \end{cases}
    \label{eq:fast_dimension}
\end{equation}
Independently of the modality, each measurement can be decomposed as
\begin{equation}
    x^{(j)}[m,k] = s^{(j)}[k] + d^{(j)}[m,k] + w^{(j)}[m,k],
    \label{eq:decomposition}
\end{equation}
where $s^{(j)}[k]$ collects the contributions of static objects, which are largely time-invariant along slow-time, $d^{(j)}[m,k]$ contains the structured temporal variations in amplitude and phase induced by human motion, and $w^{(j)}[m,k]$ is measurement noise. It is the dynamic component $d^{(j)}[m,k]$ that carries the activity information and forms the basis for activity recognition.
 
The objective is to learn, for each modality $j$, a classification function
\begin{equation}
    f\!\left(X^{(j)}\right) = A,
    \label{eq:classification}
\end{equation}
mapping the observation matrix to the activity label $A$ associated with the corresponding time window. The activity labels are defined in \Cref{sec:label_definitions}. The remainder of this section specifies the physical origin of $x^{(j)}[m,k]$ and the interpretation of the fast dimension for each sensing modality.

\subsubsection{FMCW}
\label{sec:system_model_fmcw}

An \ac{fmcw} radar continuously transmits a linearly frequency-swept signal, or \textit{chirp}, whose instantaneous frequency sweeps from $f_0$ to $f_0 + B$ over a chirp duration $T_c$ at a chirp rate $\mu = B/T_c$. The transmitted signal is:
\begin{equation}
    s_{\text{tx}}(t) = A_{\text{tx}}\,\exp\!\left(j\,2\pi\left(f_0 t
    + \frac{\mu}{2}\,t^2\right)\right).
    \label{eq:tx_signal}
\end{equation}
The echo reflected by a target at range $R$ is a delayed replica of $s_{\text{tx}}(t)$. An analog mixer multiplies the received signal with the live transmit waveform, producing a \textit{beat signal} at a single tone whose frequency is directly proportional to range:
\begin{equation}
    f_b = \frac{2\mu R}{c}.
    \label{eq:beat_freq}
\end{equation}
After low-pass filtering, the beat signal is digitised by an ADC at sampling rate $f_s$, and a quadrature demodulator yields the complex \ac{iq} samples. Chirps are transmitted repeatedly, so the chirp index takes the role of the slow-time index $m$ (with chirp repetition rate $sr_{\text{FMCW}}$), while the $N_s$ ADC samples within one chirp form the fast dimension $k$ of \Cref{eq:fast_dimension}. For a single target, the resulting measurement matrix is
\begin{equation}
    x^{(\text{FMCW})}[m,k] = A\,\exp\!\left(j\,2\pi\!\left(\frac{f_b\, k}{f_s}
    + \frac{\Delta\phi\, m}{2\pi}\right)\right) + w[m,k],
    \label{eq:iq_matrix}
\end{equation}
where $\Delta\phi = 4\pi v T_r/\lambda$ is the inter-chirp Doppler phase shift encoding radial velocity $v$, and $w[m,k] \sim \mathcal{CN}(0,\sigma^2)$ is additive complex Gaussian noise, matching the noise term of \Cref{eq:decomposition}. A static target ($v = 0$) yields a constant phase progression along $k$ only, whereas a moving person additionally modulates the phase along $m$. A 2D-FFT applied along the fast-time and slow-time dimensions of $x^{(\text{FMCW})}[m,k]$ yields the range-Doppler map used for target detection and parameter estimation.

\subsubsection{IR-UWB}
\label{sec:system_model_ir_uwb}

\Ac{uwb} transmits over an absolute bandwidth exceeding 500~MHz, enabling extremely short time-domain pulses. A 500~MHz system produces pulses of approximately 2~ns, yielding finer ranging resolution than narrowband technologies.

An \ac{ir-uwb} radar periodically transmits these short pulses and observes the resulting reflections from objects in the environment. Reflections from more distant objects arrive later, each distinct propagation path constituting a multipath component. The superposition of these delayed reflections forms the \ac{cir},
\begin{equation}
    \mathrm{CIR}(t) = \sum_{p} A_p\,\delta(t - \tau_p) + n(t),
    \label{eq:cir}
\end{equation}
where $A_p$ and $\tau_p$ are the amplitude and delay of path $p$, and $n(t)$ is noise. Static objects produce constant multipath components over time, while a moving person introduces time-varying changes in amplitude and phase, in accordance with the decomposition of \Cref{eq:decomposition}.

Successive pulse transmissions produce a sequence of \acp{cir} that capture the temporal evolution of the propagation channel. Each transmitted and successfully received \ac{uwb} packet yields one \ac{cir}, such that the slow-time axis corresponds to successive packet transmissions. Packets are transmitted at regular intervals on the order of milliseconds, resulting in a slow-time sampling rate $sr_{\text{IR-UWB}}$ as defined in \Cref{eq:slow_time_rows}, which is directly determined by the packet transmission frequency. For each received packet $m$, the reflected signal components are sampled as complex \ac{iq} values at nanosecond resolution, forming the fast-time samples. This fast-time resolution is determined by the timing resolution of the \ac{uwb} radio and defines the granularity with which multipath components can be resolved in delay. The $k$-th fast-time sample corresponds to delay bin $\tau_k$, so that
\begin{equation}
    x^{(\text{IR-UWB})}[m,k] = \mathrm{CIR}_m(\tau_k),
    \qquad 1 \leq k \leq N_f,
    \label{form:cir_matrix}
\end{equation}
where $\mathrm{CIR}_m$ denotes the \ac{cir} measured for the $m$-th received packet. Each row of $X^{(\text{IR-UWB})}$ thus represents a single packet-level \ac{cir}. The slow-time axis captures the evolution of the channel across consecutive packet transmissions, while the fast-time axis maps directly to propagation delay and, consequently, range.

\subsubsection{Wi-Fi}
\label{sec:system_model_wi_fi}

Wi-Fi is a communication-first technology that transmits \ac{ofdm} waveforms over multiple narrowband subcarriers. While designed for communication, the received packets also provide channel estimates that can be exploited for sensing. Each received packet provides one channel estimate per subcarrier, so the packet index takes the role of the slow-time index $m$ (with packet rate $sr_{\text{Wi-Fi}}$), and the $N_c$ subcarriers form the fast dimension $k$. The key sensing signal is the per-subcarrier \ac{csi},
\begin{equation}
    x^{(\text{Wi-Fi})}[m,k] = \sum_{p} \alpha_p[m]\,e^{-j2\pi f_k \tau_p[m]} + w[m,k],
    \label{form:csi_matrix}
\end{equation}
where $\alpha_p[m]$ and $\tau_p[m]$ denote the complex attenuation and delay of path $p$ at slow-time $m$, $f_k$ is the frequency of subcarrier $k$, and $w[m,k]$ is measurement noise \cite{10.1145/3310194}. Paths reflecting off static objects have essentially constant $\alpha_p$ and $\tau_p$ and thus contribute to the static component $s^{(\text{Wi-Fi})}[k]$ of \Cref{eq:decomposition}, while human motion causes time-varying $\alpha_p[m]$ and $\tau_p[m]$, producing the dynamic variations $d^{(\text{Wi-Fi})}[m,k]$ in both amplitude and phase.

\section{Experimental setup}

\subsection{Measurement setup}
\label{sec:measurement_setup}

The dataset was collected in the HomeLab in Zwijnaarde~\cite{idlab_idlab_2024}, a residential test environment developed by Ghent University and imec for evaluating \ac{iot}, smart-home, and healthcare applications. Measurements were conducted in a bedroom environment representative of an assisted-living setting, containing a bed, chair, and table. As shown in \Cref{fig:system_overview}, the sensing hardware was mounted on the ceiling above the bed and remained fixed throughout the measurement campaign.

A key aspect of the measurement campaign is that \ac{fmcw}, \ac{ir-uwb}, and Wi-Fi data were recorded simultaneously to a common time reference, allowing all three modalities to observe identical activity executions. Consequently, performance differences can be attributed to the sensing modality rather than to differences in participant behaviour or activity execution.

\begin{figure}[h!]
    \centering
    \includegraphics[width=0.95\linewidth]{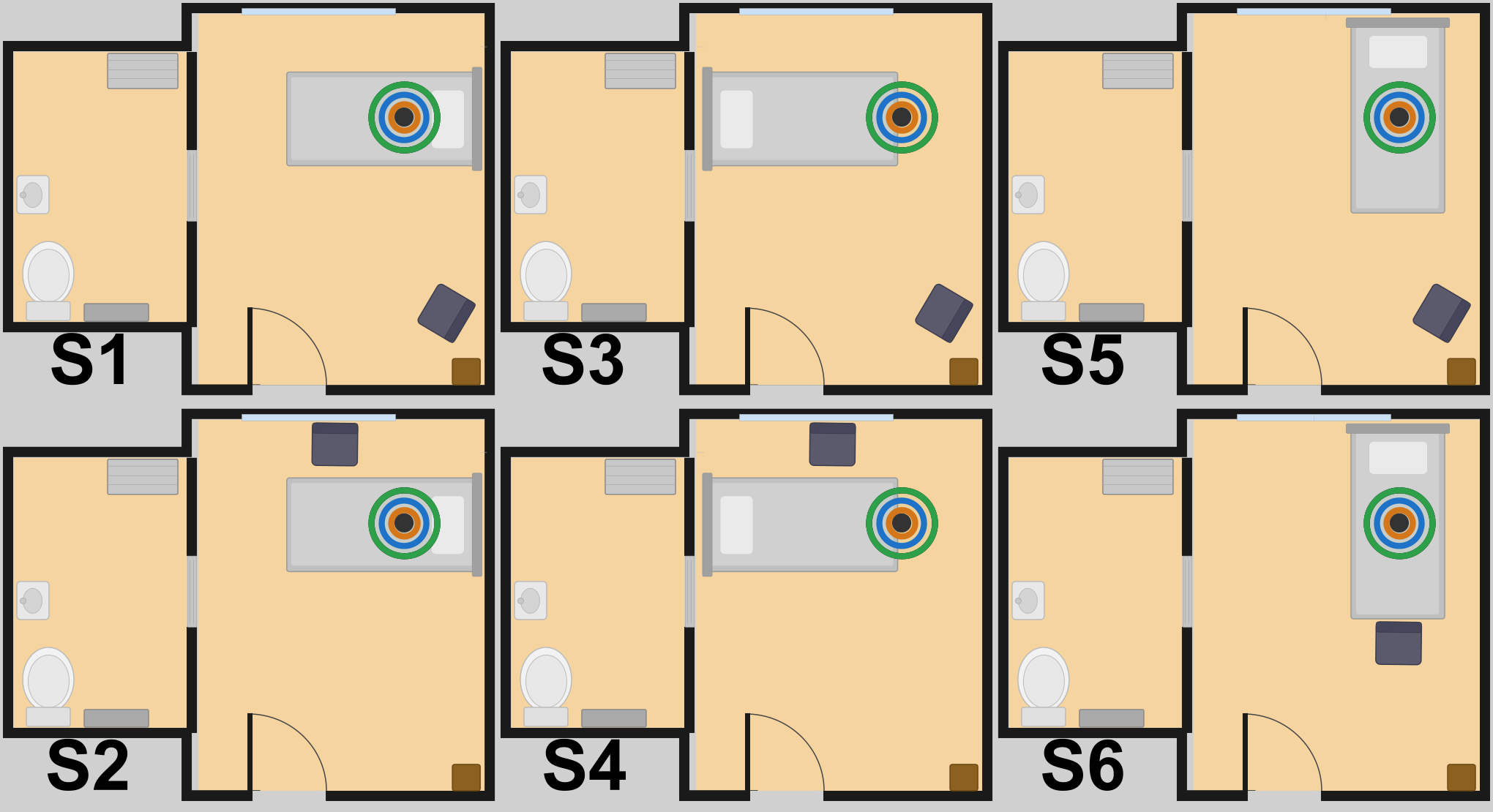}
    \caption{Overview of the six room setups used in the study. The bed and chair positions were varied to create different subject positions relative to the fixed ceiling-mounted radar setup. The radar position remained unchanged across all scenarios and is indicated by the coloured markers.}
    \label{fig:room_layouts}
\end{figure}

To evaluate robustness across environments, six room layouts were created by varying the bed and chair positions while keeping the ceiling-mounted sensing hardware fixed (\Cref{fig:room_layouts}). Layouts 1, 2, 5, and 6 place the bed beneath or close to the radar, whereas layouts 3 and 4 position it farther away or at a less favorable orientation, creating more challenging conditions for in-bed activity recognition. Seated activities are similarly more challenging when the chair is farther from the radar or positioned at a wider angle relative to the sensing direction.

A total of 20 participants (14 male and 6 female, aged 21–67 years, height: 159–189 cm, weight: 50–92 kg) completed the protocol in six room layouts, yielding up to 120 person-scenario recordings. The protocol and consent procedure were approved by the relevant ethics committees of KU Leuven (SMEC; G-2024-8332) and Ghent University, and all participants provided written informed consent.

Activity execution was guided by \ac{tts} instructions describing realistic actions, such as \textit{Walk to the left side of the bed}, \textit{Sit down on the edge of the bed}, or \textit{Go to the bathroom and wait there for a moment}. Unlike protocols that instruct participants to repeatedly perform isolated activities, these instructions encouraged participants to act within a natural context rather than repeatedly executing isolated movements. Participants were free to perform each action using their own gait, posture, and speed, resulting in realistic variability in both activities and transitions. This still leaves the timing of each action externally cued rather than fully spontaneous, but preserves natural movement transitions that repeated, isolated actions do not capture.

\subsection{Hardware}
\label{sec:hardware}

The \ac{fmcw} data was recorded with a Texas Instruments IWR6843AOP radar module operating in the 60-64~GHz band. The device integrates all sensing hardware on a single chip and provides three transmit and four receive antennas in a monostatic configuration, yielding $N_r = 12$ virtual channels. Within each 50~ms frame, 96 chirp loops are transmitted per antenna per frame which determines the unambiguous Doppler range. Its main configuration parameters are listed in \Cref{tab:sensing_config}.

The \ac{ir-uwb} data was collected with the Qorvo QM33120WDK1, a low-cost \ac{cots} development kit based on the DW3000 transceiver, operating in UWB channel~5 (centered at 6.49~GHz with 499.2~MHz bandwidth). The setup uses separate transmitter and receiver nodes, each consisting of a Nordic nRF52840 \ac{dk} and a DW3000 shield, placed approximately 30~cm apart. Despite this physical separation, the small baseline relative to the monitored area allows the setup to be treated as pseudo-monostatic. In contrast to the \ac{fmcw} radar and Wi-Fi platform, the two \ac{ir-uwb} nodes are controlled by separate boards without a shared clock backbone, which introduces a time-varying phase drift \cite{de_moerloose_towards_2026} corrected in the preprocessing of \Cref{eq:phase_correction}. The transmitter uses an omnidirectional antenna, while the receiver uses a directional antenna to increase sensitivity towards the monitored area. One \ac{cir} is acquired every 6.67~ms, which sets the slow-time rate $sr_{\text{IR-UWB}} = 150$~Hz. The corresponding configuration parameters are listed in \Cref{tab:sensing_config}.

The Wi-Fi \ac{csi} data was collected with a ZedBoard \ac{fpga} development kit equipped with an AD-FMCOMMS2-EBZ \ac{rf} front-end~\cite{jiao_single-input-multiple-output_2025}. The platform runs openwifi~\cite{Xianjun_openwifi_2020}, an open source \ac{sdr} implementation of IEEE 802.11a/g/n that exposes \ac{csi} from the physical-layer channel estimator. The system operates in a monostatic configuration with a shared transmitter-receiver clock, eliminating carrier frequency offset, sampling frequency offset, and symbol timing offset. Packets are transmitted at a mean interval of 2~ms, corresponding to a mean slow-time rate of $sr_{\text{Wi-Fi}} = 500$~Hz. The remaining configuration parameters are listed in \Cref{tab:sensing_config}.

The hardware-determined performance parameters in \Cref{tab:sensing_config} follow directly from these configurations. The range resolution is $\Delta R = c/(2B)$, so the bandwidth ratio of roughly $2{:}1{:}0.04$ between \ac{fmcw}, \ac{ir-uwb}, and Wi-Fi translates into range resolutions of 0.126~m, 0.30~m, and 7.5~m: \ac{fmcw} resolves the body into multiple range cells, \ac{ir-uwb} into a few, and Wi-Fi's single range bin exceeds the room dimensions. The maximum unambiguous velocity is $v_{\mathrm{max}} = \lambda/(4\,T_{\mathrm{slow}})$, where $T_{\mathrm{slow}}$ is the slow-time sampling interval: the chirp repetition period for \ac{fmcw}, and the frame or packet interval for \ac{ir-uwb} and Wi-Fi. All three comfortably exceed the velocities of typical in-room human motion. The Doppler resolution, by contrast, is not a hardware constant but is determined by the coherent observation interval used for Doppler processing, $\Delta v = \lambda/(2\,T_{\mathrm{obs}})$, and the corresponding values are therefore established in \Cref{sec:preprocessing}. They are nevertheless included in \Cref{tab:sensing_config} to keep all comparison parameters in a single reference table.




\begin{table}[t]
\caption{Configuration and performance parameters of the three sensing modalities.}
\label{tab:sensing_config}
\centering
\begin{tabular*}{\columnwidth}{@{\extracolsep{\fill}}l|lll}
\toprule
\textbf{Parameter} & \textbf{FMCW} & \textbf{IR-UWB} & \textbf{Wi-Fi} \\
\midrule
\multicolumn{4}{l}{\textit{Antenna setup}} \\[1pt]
\quad TX antennas             & 3              & 1             & 1 \\
\quad RX antennas             & 4              & 1             & 1 \\
\addlinespace[5pt]
\multicolumn{4}{l}{\textit{Signal}} \\[1pt]
\quad Center frequency        & 61.34\,GHz     & 6.49\,GHz     & 2.4\,GHz \\
\quad Wavelength              & 4.94\,mm       & 46.2\,mm      & 125\,mm \\
\quad Bandwidth               & 1180.05\,MHz   & 500\,MHz      & 20\,MHz \\
\quad Slope                   & 54.71\,MHz/$\mu$s & \nr        & \nr \\
\addlinespace[5pt]
\multicolumn{4}{l}{\textit{Acquisition}} \\[1pt]
\quad Frame interval          & 50\,ms         & 6.67\,ms      & 2\,ms \\
\quad Slow-time interval $T_{\mathrm{slow}}$ & 0.28\,ms & 6.67\,ms & 2\,ms \\
\quad Chirp loops             & 96             & \nr           & \nr \\
\quad Sampling frequency      & 2950\,ksps     & \nr           & \nr \\
\quad Range samples           & 64             & 50            & \nr \\
\quad Subcarriers             & \nr            & \nr           & 53 \\
\addlinespace[5pt]
\multicolumn{4}{l}{\textit{Performance}} \\[1pt]
\quad Range resolution        & 0.126~m       & 0.300~m      & 7.5~m \\
\quad Maximum range           & 6.46~m        & 7.5~m       & \nr \\
\quad Doppler resolution      & 0.093~m/s     & 0.009~m/s    & 0.156~m/s \\
\quad Maximum Doppler         & 4.471~m/s     & 1.73~m/s     & 15.6~m/s \\
\bottomrule
\end{tabular*}
\end{table}

\subsection{Label definitions}
\label{sec:label_definitions}

Each \ac{tts} instruction was mapped to an activity label during annotation. Two label sets were derived from the same recordings: a fine-grained set for detailed \ac{har} and a coarse sleep-monitoring set for deployment-oriented sleep analysis. Semantically equivalent instructions were mapped to a common activity label irrespective of location. For example, \textit{Sit down on the edge of the bed} and \textit{Sit down on the chair} are both mapped to ``SIT DOWN''. In both label sets, ``NO ACTIVITY'' denotes no relevant movement in the monitored bedroom area, including intervals where the participant is outside the room and should not produce an in-room activity signature.

The fine-grained label set contains ten classes: ``NO ACTIVITY'', ``WALK'', ``STAND UP'', ``SIT DOWN'', ``LIE ON BED'', ``GET UP BED'', ``ANXIOUS'', ``EATING'', ``WAVE HANDS'', and ``CLAP HANDS''. It preserves detailed activity categories, including walking, posture transitions, in-bed restlessness, and hand movements, to evaluate how well each sensing modality distinguishes fine and potentially subtle motion patterns.

For sleep-monitoring analysis, the activity labels were remapped into four coarse classes forming an ordinal scale of sleep disruption: ``NO ACTIVITY'', ``ANXIOUS'', ``INTERRUPTION'', and ``WANDER'', representing progressively increasing disturbance to sleep. Activities unrelated to sleep behaviour, such as eating or hand gestures, were omitted from this label set.

``NO ACTIVITY'' denotes the absence of relevant movement in the monitored bedroom. ``ANXIOUS'' captures restless in-bed motion, such as tossing or turning. ``INTERRUPTION'' represents larger posture changes, including sitting on the bed edge, standing up, or lying down again. ``WANDER'' denotes out-of-bed walking and represents the highest level of sleep disruption. This label set therefore focuses on sleep-relevant behavioural states rather than detailed activity categories.

\section{Methodology}

\subsection{Preprocessing}
\label{sec:preprocessing}

This section describes how the measurement matrices $X^{(j)}$ of \Cref{eq:general_matrix} are transformed into the image-like representations
fed to the \ac{cnn}. Selecting the preprocessing parameters is a trade-off: they should be as similar as possible across technologies so that the comparison reflects the sensing modality rather than the processing, while still not disadvantaging any modality through choices tailored to another. The parameters below aim to balance this trade-off. In particular, the coherent observation interval $T_{\mathrm{obs}}$ over which each Doppler transform is computed, one 96-chirp frame for \ac{fmcw}, the $2.5$~s window for \ac{ir-uwb}, and the $0.4$~s \ac{stft} window for Wi-Fi, determines the effective Doppler resolution $\Delta v = \lambda/(2\,T_{\mathrm{obs}})$ of each modality, as reported in \Cref{tab:sensing_config}.

\subsubsection{FMCW}

The radar acquires the measurement matrix $x^{(\text{FMCW})}[m,k]$ of \Cref{eq:iq_matrix} in frames of $N_c$ chirps and, in addition, across $N_r$ virtual antennas\footnote{A virtual antenna is defined as a transmit-receive antenna pair.}. Each ADC frame is therefore represented by $N_r$ antenna-specific matrices $\mathbf{F}_r \in \mathbb{C}^{N_c \times N_s}$, where $\mathbf{F}_r$ denotes the measurements observed by virtual antenna $r$.

A two-dimensional \ac{fft} is applied to each antenna slice: a first
transform along the fast-time axis ($N_s$ samples, index $k$) resolves
range, and a second along the slow-time axis ($N_c$ chirps, index $m$)
resolves Doppler, yielding one \ac{rdm} per antenna,
\begin{equation}
    \mathrm{RDM}_r[p,q] = \left| \mathcal{F}_{N_c}\!\big\{ \mathcal{F}_{N_s}\{ \mathbf{F}_r \} \big\} \right|,
    \qquad r = 1,\dots,N_r,
    \label{eq:rdm_fmcw}
\end{equation}
with range bin $p$ and Doppler bin $q$. Since static reflections are
time-invariant along slow-time, they concentrate in the zero-Doppler
bin, separating the static component $s^{(\text{FMCW})}[k]$ from the
motion-induced Doppler content. To suppress this static component
prior to Doppler processing, DC removal is applied to the range
profile $\mathcal{F}_{N_s}\{\mathbf{F}_r\}[k,m]$ after the first
\ac{fft}, by subtracting its mean across chirps for each range bin
$k$,
\begin{equation}
    \widehat{\mathcal{F}}_{N_s}\{\mathbf{F}_r\}[k,m] =
    \mathcal{F}_{N_s}\{\mathbf{F}_r\}[k,m] -
    \frac{1}{N_c} \sum_{m=1}^{N_c} \mathcal{F}_{N_s}\{\mathbf{F}_r\}[k,m],
    \label{eq:dc_removal}
\end{equation}
before the second \ac{fft} is applied to obtain the clutter-suppressed
\ac{rdm} in \eqref{eq:rdm_fmcw}.

The twelve TX-RX antenna pairs are summed non-coherently to improve the \ac{snr}. Taking the maximum over range for each Doppler bin then produces a one-dimensional Doppler profile without explicit position information.
 
Applying this pipeline per frame and stacking the successive Doppler profiles along the time axis yields a two-dimensional Time-Doppler map $D[t,q]$ capturing the temporal evolution of radial velocity. To build the model input, a strided windowing operation is applied to $D$ with a window length of $2.5$~s and a per-sample stride, i.e., the $50$~ms frame interval, maximising data diversity.
These steps are illustrated in Fig.~\ref{fig:system_overview}.

\subsubsection{IR-UWB}

As described in \Cref{sec:system_model_ir_uwb}, the \ac{ir-uwb} receiver outputs a raw \ac{iq} \ac{cir} at every slow-time index $m$, forming the measurement matrix $x^{(\text{IR-UWB})}[m,k]$ of \Cref{form:cir_matrix}.

Because the transmitter and receiver run on separate boards without a shared clock (\Cref{sec:hardware}), the \ac{cir} phase is corrupted by a time-varying drift. Following the phase correction approach from De~Moerloose et al.~\cite{de_moerloose_towards_2026}, the phase of the \ac{fp} sample at fast-time index $k_{\mathrm{fp}}$ is taken as the per-\ac{cir} phase offset and subtracted from every fast-time sample,
\begin{equation}
    \tilde{x}[m,k] = x^{(\text{IR-UWB})}[m,k]\;
    e^{-j \angle x^{(\text{IR-UWB})}[m, k_{\mathrm{fp}}]},
    \label{eq:phase_correction}
\end{equation}
making the phase coherent across slow-time while preserving the activity-induced phase variations.

In the received \ac{cir} signals, the useful activity information resides in the multipath components arriving beyond the \ac{fp} index. The \ac{fp} corresponds to the direct \ac{los} path between transmitter and receiver and therefore contains little information about the surrounding environment. The first 12 fast-time samples are therefore discarded, as they include both pre-\ac{fp} samples and the \ac{fp} region itself. Moreover, these early delay bins are often affected by receiver saturation and strong static components such as ceiling reflections. The remaining samples are retained as $\tilde{x}[m,k]$ for $k>12$.

A strided windowing operation is then applied along the slow-time axis using windows of $2.5$~s and a stride of $0.05$~s. A per-sample stride is avoided because the \ac{ir-uwb} slow-time rate $sr_{\text{IR-UWB}}$ is substantially higher than that of \ac{fmcw}, which would greatly inflate the dataset without adding meaningful diversity. The stride is instead matched to the \ac{fmcw} frame interval for comparability.

For each window $W$, a Doppler spectrum is computed independently for every fast-time bin using an \ac{fft} along the slow-time axis,
\begin{equation}
    \mathrm{RDM}_W[k,q] = \left| \mathcal{F}_{m \in W}\{ \tilde{x}[m,k] \} \right|,
    \label{eq:rdm_uwb}
\end{equation}
producing a Range-Doppler map with range bin $k$ and Doppler bin $q$. The Doppler axis is cropped to $[-25,25]$~Hz, identified through a per-modality grid search as containing the activity-induced motion, while frequencies outside this range were found to carry little useful information. Unlike \ac{fmcw}, the range dimension is retained because it still provides discriminative spatial information. The weaker contrast of the single-channel \ac{ir-uwb} returns makes selecting one range bin less reliable, and retaining all bins performed better than collapsing the range axis.

The overlapping Range-Doppler maps form a temporal sequence that can be viewed as a three-dimensional tensor with time, range, and Doppler dimensions. To suppress residual static clutter, a background template is estimated independently for every range-Doppler cell by taking the running median of the corresponding cell over a temporal neighbourhood of $T_{\mathrm{bg}}=10$~s,
\begin{align}
\widetilde{\mathrm{RDM}}_W[k,q]
={}&
\mathrm{RDM}_W[k,q]
\nonumber\\
&-
\operatorname{median}_{W'\in\mathcal N(W)}
\mathrm{RDM}_{W'}[k,q].
\label{eq:background_subtraction_uwb}
\end{align}
where $\mathcal{N}(W)$ denotes the temporal neighbourhood centred on $W$ spanning $T_{\mathrm{bg}}$. The median is therefore evaluated independently for every $(k,q)$ location through time, producing a slowly varying clutter estimate that is subtracted from the current Range-Doppler map while preserving transient motion. Because the temporal neighbourhood is centred on the current window, the method is non-causal and introduces up to $T_{\mathrm{bg}}/2=5$~s of latency in a real-time deployment.

\subsubsection{Wi-Fi}

For Wi-Fi, the input is the \ac{csi} measurement matrix $x^{(\text{Wi-Fi})}[m,k]$ of \Cref{form:csi_matrix}, with slow-time (packet) index $m$ and subcarrier index $k$. 

As Wi-Fi is connection-based, packets are transmitted only when permitted by the \ac{mac} layer, resulting in irregular slow-time sampling. Since the resulting timing jitter strongly affects the sensitive phase, the complex \ac{csi} samples are first interpolated onto a uniform slow-time grid at the mean packet rate of $500$~Hz, yielding the uniformly sampled matrix $x_{\mathrm{u}}[m,k]$. 

Each of the $N_c=53$ subcarriers is subsequently mapped to the Doppler domain using an \ac{stft} with a window length of $200$ samples ($0.4$~s) and an overlap of $175$ samples ($0.35$~s). The resulting per-subcarrier Time-Doppler maps $S_k[t,q]$ are averaged to form a single two-dimensional representation, 
\begin{equation} S[t,q] = \frac{1}{N_c} \sum_{k=1}^{N_c} \left|S_k[t,q]\right|, \label{eq:subcarrier_averaging} \end{equation} 
where $t$ denotes the STFT time index and $q$ the Doppler bin. The \ac{stft} yields a Doppler spectrum up to $250$~Hz, which is cropped to $[-80,80]$~Hz, identified through a per-modality grid search as containing the activity-induced motion. 

The resulting Time-Doppler maps form a temporal sequence. To suppress residual static clutter, a running median is evaluated independently for every Doppler bin over a temporal neighbourhood of $T_{\mathrm{bg}}=10$~s and subtracted from the current Time-Doppler map,
\begin{equation}
\widetilde{S}[t,q]
=
S[t,q]
-
\operatorname{median}_{t'\in\mathcal{N}(t)}
S[t',q],
\label{eq:background_subtraction}
\end{equation}
where $\mathcal{N}(t)$ denotes the temporal neighbourhood centred on $t$ spanning $T_{\mathrm{bg}}$. As for \ac{ir-uwb}, the running median is evaluated independently for every Doppler bin through time, suppressing slowly varying clutter while preserving transient motion. The same non-causality and corresponding latency therefore apply.

The clutter-suppressed Time-Doppler map is finally segmented into samples using a strided windowing operation with windows of $2.5$~s and a stride of $0.05$~s. As for \ac{ir-uwb}, a per-sample stride is avoided because it would unnecessarily inflate the dataset. Instead, the stride is matched to the \ac{fmcw} frame interval for comparability across modalities.

\subsection{CNN based classification}

A uniform \ac{cnn} architecture is applied to all three modalities, so that classification differences reflect the sensing technology rather than the classifier. Each modality produces a 2D input map of shape $(T \times F)$: for \ac{fmcw} and Wi-Fi the axes correspond to time and Doppler, for \ac{ir-uwb} they correspond to range and Doppler.

The network consists of five convolutional blocks followed by two fully connected layers, as illustrated in \Cref{tab:cnn}. Each convolutional block applies a convolution with same-padding ($3{\times}3$ in the first three blocks, $2{\times}2$ in the last two), followed by \ac{elu} activation and batch normalisation. The number of filters increases as 8, 16, 32, 64, 64. The first block applies $1{\times}2$ max-pooling, down-sampling only the feature axis $F$. The remaining four blocks apply $2{\times}2$ pooling, down-sampling both axes. After flattening, two dense layers of size 32 precede a softmax output layer of size $N_c$.
 
The architecture and training hyperparameters were selected by grid search on the training data using the same \ac{lopo} cross-validation strategy as in the final evaluation, while keeping the test fold unseen. The selected architecture was fixed across modalities to ensure a fair comparison. Regularisation is applied through $L_2$ weight decay ($\lambda = 10^{-2}$) on all learnable layers, batch normalisation in the convolutional blocks, and dropout only in the classifier head. The latter is motivated by the shared-weight structure and batch normalisation of the convolutional stack, while the dense layers contain most classifier-specific parameters. Batch size, learning rate, and learning-rate schedule were tuned per modality, and class weighting is applied to compensate for class imbalance.

\begin{table}[h!]
\caption{Uniform CNN architecture applied to all three modalities. The input is a single-channel 2D map of shape $(T \times F)$, where the two axes correspond to time and Doppler for FMCW and Wi-Fi, and to range and Doppler for IR-UWB. $N_c$ denotes the number of output classes.}
\label{tab:cnn}
\centering
\begin{tabular}{|l|l|} \hline
\textbf{Stage} & \textbf{Configuration} \\\hline
Input & $(T \times F \times 1)$ \\\hline
Conv Block 1 & Conv2D(8, 3$\times$3) + ELU + BN \\
& MaxPool(1$\times$2) \\\hline
Conv Block 2 & Conv2D(16, 3$\times$3) + ELU + BN \\
& MaxPool(2$\times$2) \\\hline
Conv Block 3 & Conv2D(32, 3$\times$3) + ELU + BN \\
& MaxPool(2$\times$2) \\\hline
Conv Block 4 & Conv2D(64, 2$\times$2) + ELU + BN \\
& MaxPool(2$\times$2) \\\hline
Conv Block 5 & Conv2D(64, 2$\times$2) + ELU + BN \\
& MaxPool(2$\times$2) \\\hline
Classifier & Flatten \\
& Dense(32) + ELU + Dropout(0.25) \\
& Dense(32) + ELU \\
& Dense($N_c$) + Softmax \\\hline
\end{tabular}
\end{table}

\subsection{Generalization evaluation strategies}
\label{sec:experimental_strategies}

Three leave-$x$-out cross-validation strategies are applied to both the fine-grained \ac{har} and coarse-grained sleep-monitoring label sets. All results are reported as macro F1 score, which accounts for precision, recall, and class imbalance. Each reported value is the mean over folds, with the standard deviation indicating fold-to-fold variability.

The following cross-validation strategies are considered:
\begin{itemize}
    \item \ac{lopo}: one participant is left out per fold as validation set, yielding 20 folds. This measures generalisation to unseen people within a seen environment.
    
    \item \ac{loso}: This evaluates generalisation to unseen room layouts. However, as illustrated in \Cref{fig:room_layouts}, the protocol remains partly optimistic because leaving out one scenario still exposes either the corresponding bed position or chair position through other scenarios. Only their combination remains unseen.
    
    \item \ac{lobpo}: scenarios are left out in pairs, namely 1-2, 3-4, and 5-6. Each fold therefore withholds one complete bed configuration, making this the strictest evaluation of cross-layout generalisation.
\end{itemize}

Each strategy reports macro F1 both before and after voting, a temporal smoothing step in which all window-level predictions belonging to the same ground-truth activity are replaced by their majority-voted label. The voted score therefore reflects event-level recognition performance while reducing the influence of isolated window-level misclassifications.

\section{Results and analysis}
\label{sec:results_and_analysis}
This section analyzes the performance of the three sensing modalities under the considered evaluation protocols and provides a signal-level interpretation of the observed differences by linking them to their underlying sensing principles, signal representations, and preprocessing choices.
\Cref{tab:overview_result_per_modality_evaluation} summarises the macro F1 scores for both label sets, the three cross-validation tiers (\ac{lopo}, \ac{loso} and \ac{lobpo}), and as explained in \Cref{sec:experimental_strategies}, the values before and after voting. The results separate along two distinct axes. In absolute accuracy within a known or similar layout, \ac{ir-uwb} leads: it attains the best fine-grained \ac{lopo} score ($89.0\%$ against $83.4\%$ for \ac{fmcw}) and matches \ac{fmcw} under \ac{loso}. In robustness to an unseen room layout, \ac{fmcw} stands alone: it is the only modality whose fine-grained accuracy is essentially unchanged under \ac{lobpo}, while \ac{ir-uwb} and Wi-Fi each lose about $10$ percentage points. Wi-Fi is consistently the weakest on the fine-grained task. At the coarse granularity used for sleep monitoring, all three modalities exceed $92\%$ under every protocol, including \ac{lobpo}.

To explain these differences, the following section will first relate these outcomes to the sensing and preprocessing properties of the three modalities (\Cref{sec:results_mechanism}). The fine-grained results are then analysed across people, post-processing and room layouts (\Cref{sec:results_fine}), followed by the coarse sleep-monitoring results (\Cref{sec:results_coarse}).

\begin{table*}[h!]
\caption{Macro F1 scores (\%, mean$\pm$std over folds) for fine-grained activity recognition and coarse-grained sleep monitoring, before and after label voting, under \ac{lopo}, \ac{loso}, and \ac{lobpo} cross-validation. IR-UWB performs best on the easier cross-subject task, while FMCW is most robust to unseen room layouts. All modalities achieve above 92\% macro F1 on the coarse sleep-monitoring task after voting.}
\label{tab:overview_result_per_modality_evaluation}
\centering
\begin{tabular}{c|cccc|cccc|cccc}
\hline
Exp
& \multicolumn{4}{c|}{FMCW}
& \multicolumn{4}{c|}{IR-UWB}
& \multicolumn{4}{c}{Wi-Fi} \\

& \multicolumn{2}{c}{Fine} & \multicolumn{2}{c|}{Coarse}& \multicolumn{2}{c}{Fine} & \multicolumn{2}{c|}{Coarse}& \multicolumn{2}{c}{Fine} & \multicolumn{2}{c}{Coarse} \\

& \shortstack{Before\\Voting} & \shortstack{After\\Voting} & \shortstack{Before\\Voting} & \shortstack{After\\Voting}& \shortstack{Before\\Voting} & \shortstack{After\\Voting} & \shortstack{Before\\Voting} & \shortstack{After\\Voting}& \shortstack{Before\\Voting} & \shortstack{After\\Voting} & \shortstack{Before\\Voting} & \shortstack{After\\Voting} \\
\hline

LOPO
& 75.0$\pm$9 & 83.4$\pm$9 & 88.5$\pm$8 & 95.1$\pm$3
& 78.6$\pm$5 & 89.0$\pm$6 & 95.4$\pm$3 & 98.2$\pm$3
& 65.0$\pm$4 & 79.0$\pm$7 & 90.2$\pm$3 & 96.1$\pm$5 \\
\hline

LOSO
& 77.3$\pm$4 & 86.6$\pm$1 & 89.9$\pm$4 & 94.9$\pm$2
& 76.3$\pm$5 & 86.4$\pm$5 & 95.3$\pm$2 & 97.7$\pm$3
& 63.2$\pm$3 & 75.3$\pm$3 & 89.4$\pm$2 & 95.1$\pm$2 \\
\hline

LOBPO
& 75.2$\pm$2 & 83.8$\pm$2 & 88.7$\pm$4 & 93.4$\pm$2
& 68.3$\pm$4 & 78.5$\pm$4 & 90.4$\pm$5 & 94.2$\pm$6
& 57.2$\pm$0 & 68.8$\pm$0 & 85.5$\pm$3 & 92.6$\pm$2 \\
\hline

\end{tabular}
\end{table*}

\subsection{Interpretation framework}
\label{sec:results_mechanism}

The observed performance differences are driven by a fundamental trade-off between activity discriminability and layout robustness. Technologies that preserve more spatial and motion detail provide the classifier with richer information to distinguish similar activities, but also increase the risk of learning environment-specific characteristics that do not generalize to unseen room layouts. Conversely, representations that discard explicit spatial information may lose some discriminative power, yet tend to be more robust to environmental changes.

Four properties from \Cref{tab:sensing_config} are central to this trade-off. First, finer range resolution separates reflections from different body regions and from surrounding objects before the range information is combined. Second, non-coherent integration across multiple TX-RX antenna pairs increases the motion energy relative to uncorrelated noise and improves the effective \ac{snr}. Third, finer Doppler resolution separates more closely spaced radial-velocity components, providing a more detailed description of the motion. Finally, retaining range in the classifier input provides explicit spatial information, whereas collapsing it reduces the model's direct access to the subject's position. 

\ac{fmcw} combines $0.126$~m range resolution with non-coherent integration over twelve TX-RX antenna pairs. The fine range bins separate moving body returns from much of the surrounding clutter, while summing the pair-specific Range-Doppler magnitudes increases their contrast. After static suppression, the range maximum can therefore retain the strongest moving return in each Doppler bin and discard its range coordinate. Stacking these profiles preserves the combined micro-Doppler pattern while removing most explicit position information. This should favour transfer across layouts, although it discards spatial cues that could help distinguish location-bound transitions.

\ac{ir-uwb} has coarser $0.30$~m range bins and only one receive channel. Its moving returns are therefore less cleanly isolated in individual range bins, making a range-maximum collapse more likely to select clutter or only part of the distributed activity signature. Retaining the full Range-Doppler map avoids this early selection and empirically outperformed the collapsed representation. Together with its finer Doppler resolution ($0.009$ versus $0.093$~m/s for \ac{fmcw}), this provides the classifier with useful spatial structure and finer velocity detail. These cues should improve recognition of posture transitions in known layouts, but can also tie those distinctions to the ranges observed during training.

Wi-Fi provides stable \ac{csi} phase because its transmitter and receiver share a clock, so human motion is clearly visible without the phase correction required by \ac{ir-uwb}. The difficulty is separating the sources of that motion. Its nominal $7.5$~m range resolution exceeds the room dimensions, so reflections from the torso, arms, legs and different propagation paths are mixed in the channel before processing. Human motion affects the $53$ subcarriers differently, providing some diversity across frequency. However, because they span only $20$~MHz, this diversity is too limited to compensate for the missing range separation, which makes similar activities difficult to distinguish. Moreover, \ac{csi} contains the combined propagation paths of the room. The network can therefore learn layout-specific channel patterns together with the activity, causing it to overfit partly to the room layout even though the representation contains no explicit range axis.

These properties provide the framework for interpreting the aggregate and per-class results below. Because the hardware and representations were not varied independently, their individual causal contributions cannot be quantified; agreement between the sensing properties, fold-level behavior and class-specific confusions nevertheless supports the resulting modality trade-offs.

\subsection{Fine-grained activity recognition}
\label{sec:results_fine}

\subsubsection{Generalization to unseen people and effect of voting}

Per-participant fine-grained \ac{lopo} scores are listed in
\Cref{tab:per_person_accuracy}. \ac{ir-uwb} achieves the highest mean macro F1 both before and after voting ($78.6\%$ and $89.0\%$), followed by \ac{fmcw} ($75.0\%$ and $83.4\%$) and Wi-Fi ($65.0\%$ and $79.0\%$). Before voting, \ac{ir-uwb} also varies less across participants (standard deviation $5$~pp) than \ac{fmcw} ($9$~pp). Within this sample, none of the modalities shows a systematic dependence on height, weight, age or gender. This suggests that recognition is not tied to a single body profile and generalises across the participant diversity represented in the dataset, although the sample size and demographic imbalance do not allow broader conclusions about demographic robustness.

Voting improves all three modalities: Wi-Fi gains $14.0$~pp, \ac{ir-uwb} $10.4$~pp and \ac{fmcw} $8.4$~pp. The larger gains for Wi-Fi and \ac{ir-uwb} show that many window-level errors are not sustained over an entire event. After voting, however, Wi-Fi remains $4.4$~pp below \ac{fmcw} and $10.0$~pp below \ac{ir-uwb}. Because voting groups predictions using the ground-truth activity boundaries, these values quantify event-level recognition under known segmentation. A deployment without such boundaries may obtain smaller gains.

\begin{table}[t!]
\centering
\caption{Per-person leave-one-person-out (LOPO) macro F1 (\%) for fine-grained activity recognition before and after label voting, across all three sensing technologies. Demographics: height (cm), weight (kg), age, gender.}
\label{tab:per_person_accuracy}
\setlength{\tabcolsep}{3pt}
\resizebox{\columnwidth}{!}{%
\begin{tabular}{@{}rrrrcrrrrrr@{}}
\hline
 & & & & & \multicolumn{2}{c}{FMCW} & \multicolumn{2}{c}{IR-UWB} & \multicolumn{2}{c}{Wi-Fi}\\
\cline{6-7}\cline{8-9}\cline{10-11}
ID & Height & Weight & Age & Sex & \shortstack{Before\\voting} & \shortstack{After\\voting} & \shortstack{Before\\voting} & \shortstack{After\\voting} & \shortstack{Before\\voting} & \shortstack{After\\voting}\\
\hline
1  & 179 & 75 & 30 & M & \textcolor{green}{\underline{88.5}} & 90.7 & 82.3 & 89.5 & 69.1 & 91.6\\
2  & 189 & 80 & 29 & M & 79.4 & 82.6 & 76.8 & 85.1 & 66.7 & 77.6\\
3  & 173 & 82 & 60 & M & 84.7 & 92.3 & 77.9 & 86.3 & 63.2 & \textcolor{red}{\underline{68.7}}\\
4  & 159 & 50 & 40 & F & 71.2 & 82.1 & 72.7 & 87.8 & \textcolor{red}{\underline{52.6}} & 72.1\\
5  & 182 & 81 & 52 & M & 68.7 & 75.5 & 77.7 & 84.7 & 63.1 & 73.5\\
6  & 172 & 71 & 27 & M & 67.8 & 73.3 & \textcolor{red}{\underline{67.4}} & \textcolor{red}{\underline{76.1}} & 60.8 & 70.1\\
7  & 178 & 92 & 26 & M & 56.9 & \textcolor{red}{\underline{60.9}} & 67.8 & 80.4 & 64.4 & 75.9\\
8  & 182 & 73 & 22 & M & 67.4 & 88.2 & 81.4 & 94.5 & 69.4 & 84.0\\
9  & 166 & 58 & 26 & F & \textcolor{red}{\underline{56.2}} & 78.4 & 83.6 & 94.7 & 61.5 & 78.8\\
10 & 178 & 75 & 27 & M & 78.4 & 85.4 & 82.5 & \textcolor{green}{\underline{98.3}} & 69.4 & 85.7\\
11 & 187 & 72 & 21 & M & 81.9 & 78.8 & 77.8 & 83.1 & 68.7 & 76.6\\
12 & 178 & 78 & 67 & M & 68.6 & 71.7 & 81.3 & 88.9 & 59.9 & 69.2\\
13 & 168 & 66 & 27 & F & 70.9 & 84.0 & 81.9 & 91.9 & 67.6 & 77.5\\
14 & 165 & 64 & 60 & F & 79.4 & 92.0 & 78.2 & 84.1 & 65.0 & 77.0\\
15 & 172 & 62 & 27 & M & 70.3 & 72.6 & 79.0 & 88.0 & 60.0 & 73.5\\
16 & 178 & 87 & 41 & M & 88.4 & 96.6 & 83.0 & 96.6 & 65.8 & 82.2\\
17 & 174 & 65 & 27 & M & 88.2 & \textcolor{green}{\underline{98.8}} & \textcolor{green}{\underline{85.8}} & 97.6 & \textcolor{green}{\underline{70.6}} & \textcolor{green}{\underline{92.7}}\\
18 & 178 & 73 & 42 & M & 76.8 & 86.2 & 82.2 & 94.9 & 69.4 & 84.2\\
19 & 172 & 88 & 25 & F & 73.1 & 83.8 & 73.9 & 83.3 & 64.9 & 82.9\\
20 & 174 & 86 & 26 & F & 82.6 & 93.2 & 78.3 & 94.7 & 67.1 & 86.4\\
\hline
Avg & & & & & 75.0 & 83.4 & 78.6 & 89.0 & 65.0 & 79.0\\
\hline
\end{tabular}%
}
\end{table}
\subsubsection{Generalization to unseen room layouts}

Under \ac{loso}, performance after voting stays close to the \ac{lopo} baseline (\Cref{tab:overview_result_per_modality_evaluation}): \ac{fmcw} improves to $86.6\%$ and \ac{ir-uwb} matches it at $86.4\%$, while Wi-Fi falls slightly to $75.3\%$. This is expected, because each scenario moves only one piece of furniture, so every bed and chair position still appears in training and only their combination is withheld (\Cref{fig:room_layouts}). In a care-home setting, where layouts are fairly uniform, this is the realistic case, and the two radars are equivalent under it.

Although \ac{fmcw} and \ac{ir-uwb} have almost identical mean \ac{loso} scores ($86.6\%$ and $86.4\%$), their fold-to-fold stability differs. Across the six folds, the difference between the highest and lowest score is only $2.3$~pp for \ac{fmcw}, with a standard deviation of $0.8$~pp. For \ac{ir-uwb}, this range is $12.8$~pp and the standard deviation is $5.4$~pp. \ac{ir-uwb} reaches $91$--$93\%$ when scenario 1, 2 or 5 is withheld, but $80$--$82\%$ for scenarios 3, 4 and 6, where the bed is farther off-centre or the chair orientation is less favourable. \ac{fmcw} therefore remains nearly unchanged regardless of which layout is unseen, whereas \ac{ir-uwb} does not, even though their mean scores are similar.

The stricter \ac{lobpo} protocol separates the modalities more clearly. Relative to \ac{lopo}, \ac{fmcw} changes from $83.4\%$ to $83.8\%$, whereas \ac{ir-uwb} and Wi-Fi decrease by $10.5$ and $10.2$~pp, respectively. The \ac{ir-uwb} result also depends on which bed position is withheld: $73.3\%$ for scenarios 3-4, compared with $81.7\%$ for scenarios 1-2 and $80.4\%$ for scenarios 5-6. Wi-Fi remains near $69\%$ in all three folds. The layout-specific \ac{ir-uwb} loss indicates sensitivity to the retained range information, whereas Wi-Fi's uniform loss reflects its inability to resolve bed position at room scale.

\subsubsection{Per-class confusion analysis}

\begin{figure*}[t!]
    \centering
    \includegraphics[width=0.90\linewidth]{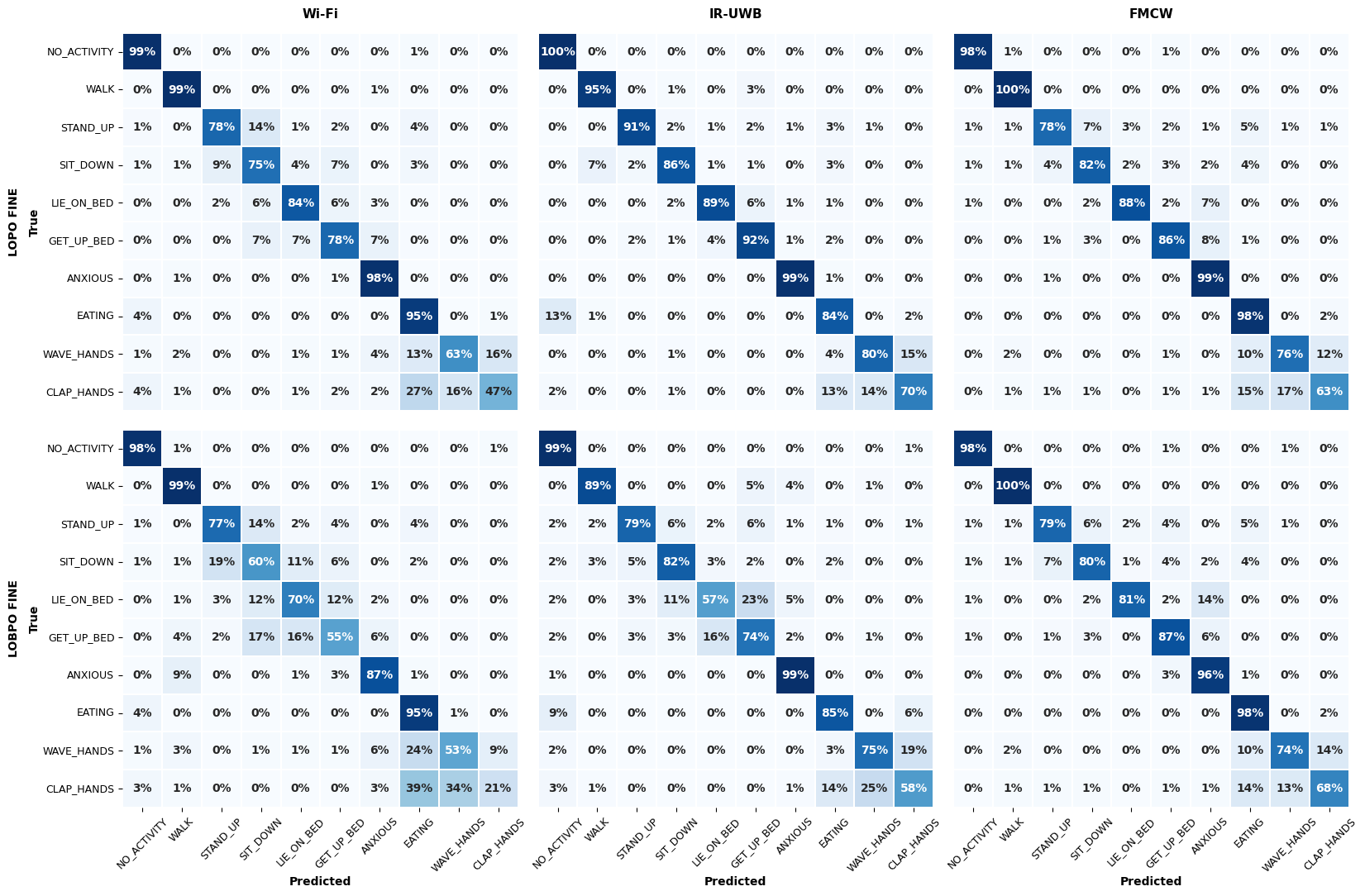}
    \caption{Confusion matrices comparing fine-grained activity classification performance across Wi-Fi, \ac{ir-uwb}, and \ac{fmcw} technologies in \ac{lopo} and \ac{lobpo} cross-validation. Rows indicate true activity labels, columns indicate predicted labels, and values are row-normalised per true class (recall, \%).}
    \label{fig:confusion_matrix_analysis_fine_grained}
\end{figure*}

The confusion matrices in \Cref{fig:confusion_matrix_analysis_fine_grained} are row-normalised and computed after voting, so each diagonal entry is per-class recall. The clearest results occur for whole-body activity: ``NO ACTIVITY'' reaches at least $98\%$ for every modality and protocol, and \ac{fmcw} recognises ``WALK'' perfectly. Under \ac{lobpo}, Wi-Fi still recalls $99\%$ of ``WALK'' but maps $9\%$ of ``ANXIOUS'' to that class, showing that movement in bed and walking can produce similar signatures when the system cannot determine where the motion occurs. \ac{ir-uwb} instead loses ``WALK'' recall, from $95\%$ to $89\%$, with errors assigned mainly to ``LIE ON BED'' and ``GET UP BED''. This shift towards bed-related classes supports the conclusion that its retained range axis ties activity recognition more closely to the positions observed during training.

Posture transitions account for much of the difference between \ac{ir-uwb} and \ac{fmcw} under \ac{lopo}. \ac{ir-uwb} exceeds \ac{fmcw} by $13$~pp on ``STAND UP'', $6$~pp on ``GET UP BED'' and $4$~pp on ``SIT DOWN''. This concentration of the advantage in related transitions is consistent with the combination of finer Doppler detail and retained range helping to distinguish their torso-motion trajectories. The present comparison cannot determine the contribution of either property separately. Under \ac{lobpo}, these classes degrade most strongly for \ac{ir-uwb} and Wi-Fi: \ac{ir-uwb} ``LIE ON BED'' falls from $89\%$ to $57\%$, with $23\%$ assigned to ``GET UP BED'', while Wi-Fi ``GET UP BED'' falls from $78\%$ to $55\%$ and is split mainly between ``SIT DOWN'' ($17\%$) and ``LIE ON BED'' ($16\%$). The concentration of errors among related transitions suggests that their distinctions transfer poorly when the bed position is unseen.

The hand-movement classes are also difficult, especially for Wi-Fi. Its ``CLAP HANDS'' recall decreases from $47\%$ to $21\%$ under \ac{lobpo}, with most errors assigned to ``EATING'' and ``WAVE HANDS''. Its unresolved channel response therefore preserves evidence of hand motion but not enough detail to separate similar gestures reliably. \ac{fmcw} is more stable on these classes and maintains $98\%$ recall for ``EATING'', although ``WAVE HANDS'' and ``CLAP HANDS'' remain mutually confused. This advantage is consistent with its finer range resolution separating body-region returns and the non-coherent sum over twelve TX-RX antenna pairs improving the \ac{snr} before the range axis is collapsed.

Finally, \ac{ir-uwb} more often maps activity to ``NO ACTIVITY'' than the other modalities. The largest example is $13\%$ of ``EATING'' under \ac{lopo}. Its lower per-bin \ac{snr}, together with residual phase instability from the unsynchronised transmitter and receiver, likely raises the motion evidence required for detection. For \ac{fmcw}, the main cross-layout change is instead ``LIE ON BED'' being assigned to ``ANXIOUS'' more often ($7\%$ to $14\%$). Once range is collapsed, both classes are represented mainly by low-velocity in-bed motion, explaining why they remain the principal \ac{fmcw} confusion.


\subsection{Coarse-grained sleep monitoring}
\label{sec:results_coarse}

The coarse label set merges posture transitions into ``INTERRUPTION'' and omits activities that do not fit in the sleep-disruption context (\Cref{sec:label_definitions}). After voting, \ac{lopo} macro F1 reaches $98.2\%$ for \ac{ir-uwb}, $96.1\%$ for Wi-Fi and $95.1\%$ for \ac{fmcw}.

Cross-layout differences are also smaller than for the fine-grained task. Under \ac{loso}, the scores are $97.7\%$, $95.1\%$ and $94.9\%$ for \ac{ir-uwb}, Wi-Fi and \ac{fmcw}, respectively. Under \ac{lobpo}, all three remain within $1.6$~pp: \ac{ir-uwb} reaches $94.2\%$, \ac{fmcw} $93.4\%$ and Wi-Fi $92.6\%$. The merged labels remove the difficult hand movements and most distinctions among posture transitions, so these values describe a less demanding classification problem rather than improved sensing fidelity.

The remaining class-level errors are modality-specific. Under \ac{lobpo}, \ac{fmcw} assigns $10\%$ of ``ANXIOUS'' to ``INTERRUPTION'', Wi-Fi assigns $5\%$ to ``WANDER'', and \ac{ir-uwb} ``WANDER'' recall falls from $97\%$ to $85\%$, with errors divided between ``ANXIOUS'' and ``INTERRUPTION''. \ac{ir-uwb} is also the only modality to assign non-zero proportions of all three activity classes to ``NO ACTIVITY'' ($1$--$2\%$). Although small, the latter errors are operationally important because they represent missed sleep disruptions rather than confusion between disruption levels.

\section{Deployment trade-offs}
\label{sec:comparison}

The preceding sections showed how the three sensing modalities differ in recognition accuracy and robustness. For real-world single-anchor deployment, however, the preferred modality also depends on constraints beyond \ac{har} performance. This section translates the \ac{ml} results into deployment-oriented trade-offs by considering localisation, power, cost, and communication capabilities.

\subsection{Localization performance}

Localization complements \ac{har} in real-world deployments: mapping activity hotspots provides spatial context and can improve recognition accuracy. The three modalities differ substantially in this capability. \ac{fmcw} resolves range at $0.126$~m (\Cref{tab:sensing_config}) and angle via its twelve virtual channels, enabling accurate 3D localization within the room \cite{Zhao_fmcw_tracking_2021}. The \ac{ir-uwb} kits support ranging and \ac{aoa} estimation, but their accuracy is bounded by the coarser $0.30$~m range resolution. Moreover, ranging on humans with \ac{ir-uwb} is challenging due to the body's low radar reflectivity \cite{Herbruggen_UWB_ranging_2023}, a limitation amplified in this work by the reduced radar cross-section of the ceiling-mounted geometry. Wi-Fi can in principle estimate range by transforming \ac{csi} into \ac{cir}s, but its $20$~MHz bandwidth corresponds to a single $7.5$~m range bin that exceeds the room dimensions. Although phase tracking or wider channels could improve ranging for both \ac{ir-uwb} and Wi-Fi \cite{de_moerloose_towards_2026}, \ac{fmcw} is currently the only modality of the three that provides robust 3D localization without additional infrastructure or processing.

\subsection{Power Consumption}

The three modalities differ by more than an order of magnitude in power draw. The \ac{ir-uwb} front-end is by far the least power intensive: the \ac{cots} DW3000-based transmitter-receiver system consumes on the order of $0.2$~W in continuous operation \cite{Lambrecht_low_cost_2026}, and can be driven substantially lower through device-level duty-cycling. The single-chip \ac{fmcw} sensor is significantly more demanding: the IWR6843 typically draws $1.2$-$1.75$~W \cite{TI_IWR6843}, scaling with the number of active transmit channels and the frame duty cycle, with the upper part of this range applicable to the three-transmitter configuration used in this work. Unlike \ac{ir-uwb} and \ac{fmcw}, monostatic Wi-Fi sensing cannot currently be realized with standard \ac{cots} Wi-Fi chipsets, as they do not expose the reflected-signal \ac{csi} in the presence of simultaneous transmission. Consequently, specialized \ac{sdr}-based platforms are typically required for single-device Wi-Fi radar. As a result, the Wi-Fi \ac{csi} platform consumes the most, at $3.5$-$4.0$~W (measured at $5.2$~V, $0.68$-$0.76$~A in our setup). This figure reflects the \ac{sdr} platform (ZedBoard with AD-FMCOMMS2) required for \ac{csi} access, which is not power-optimised and is not representative of a commodity Wi-Fi radio.

\subsection{Cost-effectiveness}

Hardware prices separate the modalities by an order of magnitude.\footnote{Unit prices at 1000-unit quantity from the respective distributors (Mouser, Texas Instruments, CrowdSupply), accessed June~2026.} The \ac{ir-uwb} radar is the cheapest: at roughly EUR~7 per DW3000 transceiver, even the two-node pseudo-monostatic setup totals approximately EUR~14. The single-chip \ac{fmcw} radar is in the same class at roughly EUR~20. Wi-Fi is far more expensive: to the best of our knowledge, no commercial Wi-Fi chipset currently supports the full-duplex monostatic radar mode used here, so a specialised \ac{sdr} platform is required, the cheapest we identified costing approximately EUR~320. This price is development-grade and would likely fall in a dedicated deployment, but at present the need for specialised hardware makes Wi-Fi the least cost-effective option for monostatic deployments.

\subsection{Communication capability}

Finally, the modalities differ in how naturally sensing integrates with communication. Wi-Fi is a communication standard first, so sensing reuses an existing data link at no additional spectral cost. \ac{ir-uwb} is IEEE~802.15.4z-compliant and likewise combines ranging and data communication on the same device. \ac{fmcw}, by contrast, is a sensing-only modality with no native communication and would require a separate link for data transport. Wi-Fi and \ac{ir-uwb} are therefore attractive where a single device must both sense and communicate.

\section{Conclusion}

This paper compared three ceiling-mounted \ac{cots} \ac{rf} technologies (\ac{fmcw}, \ac{ir-uwb}, Wi-Fi) for \ac{har} and sleep monitoring using a uniform \ac{cnn} on a 20-participant dataset.

In the \ac{lopo} evaluation, fine-grained recognition is led by \ac{ir-uwb} ($89.0\%$), followed by \ac{fmcw} ($83.4\%$) and Wi-Fi ($79.0\%$). All three exceed $95\%$ macro F1 for coarse sleep monitoring. Under the stricter \ac{lobpo} protocol, \ac{fmcw} remains stable at $83.8\%$, whereas \ac{ir-uwb} and Wi-Fi decrease to $78.5\%$ and $68.8\%$, respectively. \ac{ir-uwb}'s retained range and finer Doppler information favour detailed recognition in known layouts but increase position dependence, whereas \ac{fmcw}'s collapsed-range representation trades some of that detail for stronger cross-layout robustness. Wi-Fi's unresolved, multipath-dependent channel response limits both fine-grained accuracy and transfer to unseen layouts.

Deployment trade-offs extend beyond accuracy. \ac{ir-uwb} is the most cost- and power-efficient option and natively supports unified sensing and communication. \ac{fmcw} maximizes sensing robustness and cross-environment generalisation, but consumes more power and requires a separate data link. Despite the appeal of reusing existing infrastructure, Wi-Fi is currently the least practical for monostatic radar, as it requires an expensive, specialized software-defined-radio platform. Consequently, the preferred modality depends on the deployment priority: robustness favours \ac{fmcw}, whereas cost, power efficiency, and integrated connectivity favour \ac{ir-uwb}.

\section*{Declaration on the Use of Generative AI}

During the preparation of this manuscript, the authors used generative \ac{ai} tools, primarily Microsoft Copilot, to assist with language editing, paraphrasing, rewording, grammar and spelling checks, and the refinement of selected sections of the text. Microsoft Copilot was also used to assist in the development of portions of the data-processing and machine-learning code used during the study. All \ac{ai}-generated suggestions, text, and code were carefully reviewed, validated, and modified where necessary by the authors. The authors take full responsibility for the accuracy, integrity, and final content of the manuscript, including all analyses, results, and conclusions.



\bibliographystyle{IEEEtran}
\bibliography{reference.bib}

\begin{thebibliography}{10}
\providecommand{\url}[1]{#1}
\csname url@samestyle\endcsname
\providecommand{\newblock}{\relax}
\providecommand{\bibinfo}[2]{#2}
\providecommand{\BIBentrySTDinterwordspacing}{\spaceskip=0pt\relax}
\providecommand{\BIBentryALTinterwordstretchfactor}{4}
\providecommand{\BIBentryALTinterwordspacing}{\spaceskip=\fontdimen2\font plus
\BIBentryALTinterwordstretchfactor\fontdimen3\font minus \fontdimen4\font\relax}
\providecommand{\BIBforeignlanguage}[2]{{%
\expandafter\ifx\csname l@#1\endcsname\relax
\typeout{** WARNING: IEEEtran.bst: No hyphenation pattern has been}%
\typeout{** loaded for the language `#1'. Using the pattern for}%
\typeout{** the default language instead.}%
\else
\language=\csname l@#1\endcsname
\fi
#2}}
\providecommand{\BIBdecl}{\relax}
\BIBdecl

\bibitem{who_ageing_2025}
{World Health Organization}, ``Ageing and health,'' \url{https://www.who.int/news-room/fact-sheets/detail/ageing-and-health}, 2025, accessed: 2026-05-11.

\bibitem{canever_worldwide_2024}
\BIBentryALTinterwordspacing
J.~B. Canever, G.~Zurman, F.~Vogel, D.~V. Sutil, J.~B.~M. Diz, A.~L. Danielewicz, B.~D.~S. Moreira, H.~I. Cimarosti, and N.~C.~P. De~Avelar, ``\BIBforeignlanguage{en}{Worldwide prevalence of sleep problems in community-dwelling older adults: {A} systematic review and meta-analysis},'' \emph{\BIBforeignlanguage{en}{Sleep Medicine}}, vol. 119, pp. 118--134, Jul. 2024. [Online]. Available: \url{https://linkinghub.elsevier.com/retrieve/pii/S1389945724001448}
\BIBentrySTDinterwordspacing

\bibitem{wrede_understanding_2023}
\BIBentryALTinterwordspacing
C.~Wrede, A.~Braakman-Jansen, and L.~Van Gemert-Pijnen, ``Understanding acceptance of contactless monitoring technology in home-based dementia care: a cross-sectional survey study among informal caregivers,'' \emph{Frontiers in Digital Health}, vol.~5, p. 1257009, Oct. 2023. [Online]. Available: \url{https://www.frontiersin.org/articles/10.3389/fdgth.2023.1257009/full}
\BIBentrySTDinterwordspacing

\bibitem{lu_unobtrusive_2023}
\BIBentryALTinterwordspacing
W.~Lu, S.~Kumar, M.~Sandhu, and Q.~Zhang, ``An {Unobtrusive} {Fall} {Detection} {System} {Using} {Ceiling}-mounted {Ultra}-wideband {Radar},'' in \emph{2023 45th {Annual} {International} {Conference} of the {IEEE} {Engineering} in {Medicine} \& {Biology} {Society} ({EMBC})}.\hskip 1em plus 0.5em minus 0.4em\relax Sydney, Australia: IEEE, Jul. 2023, pp. 1--5. [Online]. Available: \url{https://ieeexplore.ieee.org/document/10341081/}
\BIBentrySTDinterwordspacing

\bibitem{lu_office_2025}
\BIBentryALTinterwordspacing
W.~Lu, C.~Bird, M.~Sandhu, and D.~Silvera-Tawil, ``\BIBforeignlanguage{en}{Office {Posture} {Detection} {Using} {Ceiling}-{Mounted} {Ultra}-{Wideband} {Radar} and {Attention}-{Based} {Modality} {Fusion}},'' \emph{\BIBforeignlanguage{en}{Sensors}}, vol.~25, no.~16, p. 5164, Aug. 2025. [Online]. Available: \url{https://www.mdpi.com/1424-8220/25/16/5164}
\BIBentrySTDinterwordspacing

\bibitem{guo_human_2024}
\BIBentryALTinterwordspacing
W.~Guo, S.~Yamagishi, and L.~Jing, ``Human {Activity} {Recognition} via {Wi}-{Fi} and {Inertial} {Sensors} {With} {Machine} {Learning},'' \emph{IEEE Access}, vol.~12, pp. 18\,821--18\,836, 2024. [Online]. Available: \url{https://ieeexplore.ieee.org/document/10418123/}
\BIBentrySTDinterwordspacing

\bibitem{chen_wiphase_2025}
\BIBentryALTinterwordspacing
X.~Chen, C.~Li, C.~Jiang, W.~Meng, and W.~Xiao, ``{WiPhase}: {A} {Human} {Activity} {Recognition} {Approach} by {Fusing} of {Reconstructed} {WiFi} {CSI} {Phase} {Features},'' \emph{IEEE Transactions on Mobile Computing}, vol.~24, no.~1, pp. 394--406, Jan. 2025. [Online]. Available: \url{https://ieeexplore.ieee.org/document/10681250/}
\BIBentrySTDinterwordspacing

\bibitem{gurbuz_cross-frequency_2020}
\BIBentryALTinterwordspacing
S.~Z. Gurbuz, M.~M. Rahman, E.~Kurtoglu, T.~Macks, and F.~Fioranelli, ``Cross-frequency training with adversarial learning for radar micro-{Doppler} signature classification ({Rising} {Researcher}),'' in \emph{Radar {Sensor} {Technology} {XXIV}}, A.~M. Raynal and K.~I. Ranney, Eds.\hskip 1em plus 0.5em minus 0.4em\relax Online Only, United States: SPIE, May 2020, p.~16. [Online]. Available: \url{https://www.spiedigitallibrary.org/conference-proceedings-of-spie/11408/2559155/Cross-frequency-training-with-adversarial-learning-for-radar-micro-Doppler/10.1117/12.2559155.full}
\BIBentrySTDinterwordspacing

\bibitem{ding_radar_2025}
\BIBentryALTinterwordspacing
Y.~Ding, P.~Lv, R.~Liu, Y.~Peng, and M.~Ding, ``A {Radar} {System}-{Agnostic} ({RSA}) {Learning} {Architecture} for {Human} {Activity} {Recognition},'' \emph{IEEE Sensors Journal}, vol.~25, no.~10, pp. 18\,492--18\,502, May 2025. [Online]. Available: \url{https://ieeexplore.ieee.org/document/10948888/}
\BIBentrySTDinterwordspacing

\bibitem{ibrahim_radspecfusion_2025}
\BIBentryALTinterwordspacing
A.~Ibrahim, M.~Z. Khan, M.~Imran, H.~Larijani, Q.~H. Abbasi, and M.~Usman, ``\BIBforeignlanguage{en}{{RadSpecFusion}: {Dynamic} attention weighting for multi-radar human activity recognition},'' \emph{\BIBforeignlanguage{en}{Internet of Things}}, vol.~33, p. 101682, Sep. 2025. [Online]. Available: \url{https://linkinghub.elsevier.com/retrieve/pii/S2542660525001969}
\BIBentrySTDinterwordspacing

\bibitem{chen_rf-based_2023}
\BIBentryALTinterwordspacing
Z.~Chen, C.~Cai, T.~Zheng, J.~Luo, J.~Xiong, and X.~Wang, ``{RF}-{Based} {Human} {Activity} {Recognition} {Using} {Signal} {Adapted} {Convolutional} {Neural} {Network},'' \emph{IEEE Transactions on Mobile Computing}, vol.~22, no.~1, pp. 487--499, Jan. 2023. [Online]. Available: \url{https://ieeexplore.ieee.org/document/9408395/}
\BIBentrySTDinterwordspacing

\bibitem{dahal_comparison_2024}
\BIBentryALTinterwordspacing
A.~Dahal, S.~Biswas, S.~Z. Gurbuz, and A.~C. Gurbuz, ``Comparison {Between} {Wi}-{Fi}-{CSI} and {Radar}-{Based} {HAR},'' in \emph{2024 {IEEE} {Radar} {Conference} ({RadarConf24})}.\hskip 1em plus 0.5em minus 0.4em\relax Denver, CO, USA: IEEE, May 2024, pp. 1--6. [Online]. Available: \url{https://ieeexplore.ieee.org/document/10548515/}
\BIBentrySTDinterwordspacing

\bibitem{jiao_single-input-multiple-output_2025}
\BIBentryALTinterwordspacing
X.~Jiao, T.~Havinga, W.~Liu, and I.~Moerman, ``Single-{Input}-{Multiple}-{Output} {Wi}-{Fi} {Radar} for {Vital} {Signal} {Sensing} and {Device} {Tracking},'' in \emph{2025 {IEEE} 5th {International} {Symposium} on {Joint} {Communications} \&; {Sensing} (JC\&S)}.\hskip 1em plus 0.5em minus 0.4em\relax Oulu, Finland: IEEE, Jan. 2025, pp. 1--3. [Online]. Available: \url{https://ieeexplore.ieee.org/document/10880631/}
\BIBentrySTDinterwordspacing

\bibitem{kristensen_sdr-based_2025}
\BIBentryALTinterwordspacing
A.~T. Kristensen, A.~Balatsoukas-Stimming, and A.~Burg, ``An {SDR}-{Based} {Monostatic} {Wi}-{Fi} {System} with {Analog} {Self}-{Interference} {Cancellation} for {Sensing},'' in \emph{2025 {IEEE} {International} {Symposium} on {Circuits} and {Systems} ({ISCAS})}.\hskip 1em plus 0.5em minus 0.4em\relax London, United Kingdom: IEEE, May 2025, pp. 1--5. [Online]. Available: \url{https://ieeexplore.ieee.org/document/11043800/}
\BIBentrySTDinterwordspacing

\bibitem{bakhshi_respisense_2025}
\BIBentryALTinterwordspacing
G.~Bakhshi, M.~A. Rawahi, N.~Akhshatayeva, M.~E. Hassan, M.~Elsayed, Y.~Elshenawy, S.~D. Naicker, and H.~Abou-Zeid, ``{RespiSense}: {Real}-{Time} {Respiration} {Monitoring} {Using} a {Low}-{Complexity} {WiFi} {SDR} {Platform},'' in \emph{{GLOBECOM} 2025 - 2025 {IEEE} {Global} {Communications} {Conference}}.\hskip 1em plus 0.5em minus 0.4em\relax Taipei, Taiwan: IEEE, Dec. 2025, pp. 973--979. [Online]. Available: \url{https://ieeexplore.ieee.org/document/11432470/}
\BIBentrySTDinterwordspacing

\bibitem{10.1145/3310194}
\BIBentryALTinterwordspacing
Y.~Ma, G.~Zhou, and S.~Wang, ``Wifi sensing with channel state information: A survey,'' \emph{ACM Comput. Surv.}, vol.~52, no.~3, Jun. 2019. [Online]. Available: \url{https://doi.org/10.1145/3310194}
\BIBentrySTDinterwordspacing

\bibitem{idlab_idlab_2024}
\BIBentryALTinterwordspacing
{iDlab}. (2024) {iDlab} homelab. [Online]. Available: \url{https://homelab.ilabt.imec.be/}
\BIBentrySTDinterwordspacing

\bibitem{de_moerloose_towards_2026}
\BIBentryALTinterwordspacing
J.~De~Moerloose, A.~Shahid, and E.~De~Poorter, ``Towards mm-{Level} {Accurate} {UWB} {Radar}: {High}-{Accuracy} {Phase}-{Based} {Obstacle} {Detection} through {Multi}-{Channel} {Fusion},'' 2026, version Number: 1. [Online]. Available: \url{https://arxiv.org/abs/2606.16657}
\BIBentrySTDinterwordspacing

\bibitem{Xianjun_openwifi_2020}
X.~Jiao, W.~Liu, M.~Mehari, M.~Aslam, and I.~Moerman, ``openwifi: a free and open-source ieee802.11 sdr implementation on soc,'' in \emph{2020 IEEE 91st Vehicular Technology Conference (VTC2020-Spring)}, 2020, pp. 1--2.

\bibitem{Zhao_fmcw_tracking_2021}
\BIBentryALTinterwordspacing
P.~Zhao, C.~X. Lu, J.~Wang, C.~Chen, W.~Wang, N.~Trigoni, and A.~Markham, ``Human tracking and identification through a millimeter wave radar,'' \emph{Ad Hoc Networks}, vol. 116, p. 102475, 2021. [Online]. Available: \url{https://www.sciencedirect.com/science/article/pii/S1570870521000421}
\BIBentrySTDinterwordspacing

\bibitem{Herbruggen_UWB_ranging_2023}
B.~V. Herbruggen, S.~Luchie, R.~Berkvens, J.~Fontaine, and E.~D. Poorter, ``Impact of cir processing for uwb radar distance estimation with the dw1000 transceiver,'' in \emph{2023 13th International Conference on Indoor Positioning and Indoor Navigation (IPIN)}, 2023, pp. 1--7.

\bibitem{Lambrecht_low_cost_2026}
A.~Lambrecht, S.~Luchie, J.~Fontaine, B.~V. Herbruggen, A.~Shahid, and E.~De~Poorter, ``Low-cost embedded breathing rate determination using 802.15.4z ir-uwb hardware for remote healthcare,'' \emph{IEEE Sensors Journal}, pp. 1--1, 2026.

\bibitem{TI_IWR6843}
\BIBentryALTinterwordspacing
{Texas Instruments}, \emph{IWR6843, IWR6443 Single-Chip 60 to 64GHz mmWave Sensor Datasheet}, Texas Instruments, Apr. 2025, rev. F. [Online]. Available: \url{https://www.ti.com/lit/ds/symlink/iwr6843.pdf}
\BIBentrySTDinterwordspacing

\end{thebibliography}

\begin{IEEEbiographynophoto}
{Anton Lambrecht} received the M.Sc. degree in Computer Science Engineering from Ghent University, Belgium, in 2024. Shortly thereafter, in September 2024, he joined the Department of Information Technology (INTEC) at Ghent University as a researcher within the IDLab research group, where he is currently pursuing the Ph.D. degree. His research interests include RF-based sensing using Ultra-wideband technology, with a focus on healthcare applications through machine learning techniques.
\end{IEEEbiographynophoto}

\begin{IEEEbiographynophoto}
{Reda el Hail} received his Master's degree in Signal and Image Processing from the University of Rennes 1, France, and his Engineering degree in Electrical Engineering from the Mohammadia School of Engineers in Morocco. As a Ph.D. researcher in machine learning at KU Leuven in Belgium, his current research concentrates on machine learning techniques for indoor human monitoring using millimeter-wave radar technology, with a particular emphasis on enhancing model robustness towards different aspects such as unseen environments, multiple persons and open set recognition.
\end{IEEEbiographynophoto}

\begin{IEEEbiographynophoto}
{XIANJUN JIAO} received his bachelor's degree in Electrical Engineering from Nankai university in 2001 and Ph.D. in communication and information system from Peking University in 2006. Then, he worked in the industry, such as Nokia, Microsoft and Apple, on wireless systems. In 2016, he joined IDLab, a core research group of imec embedded in Ghent University and University of Antwerp, as a senior researcher. At imec, he works on real-time SDR (Software Defined Radio) platform, such as the openwifi project, open-source Wi-Fi chip design, which is used widely for research in wireless networking, sensing and security. His main interests are designing and implementing high performance/efficient PHY and MAC layers, parallel/heterogeneous computation for wireless communications.
\end{IEEEbiographynophoto}

\begin{IEEEbiographynophoto}
{PIETER CROMBEZ} received the M.S. degree in electronic engineering from KU Leuven University, in 2005, and the Ph.D. degree in electronic engineering from KU Leuven University, in 2009, with his research on low power, reconfigurable transceivers for multistandard/multimode applications. He was a Research Assistant with the ESAT-MICAS Laboratory, KU Leuven University. He joined as the Research and Development Project Lead with Televic Healthcare in 2009, where he was responsible for the wireless research for next generation nurse call products. The main focus was on localization techniques that meet the harsh specifications for the healthcare market. Nowadays, holding the position of Research Lead, he is responsible for the long term research activities within Televic Healthcare with focus on wireless communication, localization and sensing. He has been involved in several national and European funded research projects, both as a participant and a project coordinator. He is a (co)author of several publications.
\end{IEEEbiographynophoto}

\begin{IEEEbiographynophoto}
{Dominique Schreurs} (Fellow, IEEE) received the M.Sc. degree in electronic engineering and the Ph.D. degree from the University of Leuven (KU Leuven), Leuven, Belgium, in 1992 and 1997, respectively. She has been a Visiting Scientist with Agilent Technologies, Santa Rosa, CA, USA, ETH Zürich, Zürich, Switzerland, and the National Institute of Standards and Technology, Boulder, CO, USA. She is currently a Full Professor with KU Leuven, where she is also the Chair of the Leuven ICT (the Leuven Centre on Information and Communication Technology). Her current research interests include the microwave and millimeter-wave characterization and modeling of transistors, nonlinear circuits, and bioliquids, and system design for wireless communications and biomedical applications. Prof. Schreurs served as the President of the IEEE Microwave Theory and Techniques Society from 2018 to 2019. She was an IEEE MTT-S Distinguished Microwave Lecturer. She has also served as the General Chair for the Spring Automatic RF Techniques Group (ARFTG) conferences in 2007, 2012, and 2018, and the President of the ARFTG organization from 2018 to 2019. She currently serves as the TPC Chair for the European Microwave Conference and also the Conference Co-Chair for the IEEE International Microwave Biomedical Conference. She was the Editor-in-Chief of the IEEE Transactions on Microwave Theory and Techniques.
\end{IEEEbiographynophoto}

\begin{IEEEbiographynophoto}
{PETER KARSMAKERS}  In 2010 Peter Karsmakers received his Ph.D. at KU Leuven, Department of Electrical Engineering. In 2013 as a post-doc, he co-founded the ADVISE research group at Geel campus together with a few other colleagues. From 2018 he is an Assistant Professor at KU Leuven in the Computer Science Department and since 2022 he is principal investigator at Flanders Make. His research focuses on designing machine learning algorithms that consider application-specific constraints. These can for example relate to the computing platform on which the algorithm will be deployed or background knowledge about the machine learning task.
\end{IEEEbiographynophoto}

\begin{IEEEbiographynophoto}
{Adnan Shahid} (Senior Member, IEEE) received the Ph.D. degree in Information and Communication Engineering from Sejong University, South Korea, in 2015. He is currently an Assistant Professor with the Internet Technology and Data Science Laboratory (IDLab), Ghent University–imec, where he leads the “AI/ML for Wireless” research within the IDLab Intelligent Wireless Networking (iWINe) group. He actively contributes to international standardization efforts, including the IEEE P1900.8 Working Group on Training, Testing, and Evaluating Machine-Learned Spectrum Awareness Models and the ATIS Working Group on Generative AI in Telecommunications. He has participated in several major research initiatives, including the DARPA Spectrum Collaboration Challenge (SC2) and multiple H2020, Horizon Europe, and ESA projects. He has authored over 150 publications in leading conferences and journals. His research interests include wireless physical-layer foundation models, decentralized and federated learning, radio resource management, 5G/6G networks, and non-terrestrial networks (NTN).
\end{IEEEbiographynophoto}

\begin{IEEEbiographynophoto}
{Eli De Poorter} is professor at the IDLab research group from imec \& Ghent University (https://idlab.technology). His team performs research on wireless communication technologies such as (indoor) localization solutions, connected healthcare, wireless IoT solutions and machine learning for wireless systems. He performs both fundamental and applied research. He currently has around 350 publications and has a h-index of 50 (google scholar) with around 1400 citations per year. Since 2021, Prof. De Poorter has been included yearly in the Stanford University (Stanford, CA, USA), World’s Top 2\% Scientists in the domain of Networking \& Telecommunications.
\end{IEEEbiographynophoto}

\EOD

\end{document}